%% file: main.tex
\documentclass[sigconf]{acmart}
\AtBeginDocument{%
  }

\usepackage{pifont}
\usepackage{colortbl}
\usepackage{multirow}
\usepackage{url}
\usepackage{float}

\usepackage{amsmath,amssymb,amsfonts, bm}
\usepackage{colortbl} 
\usepackage{xcolor}
\usepackage{array}  
\usepackage{subfigure}
\setcopyright{acmcopyright}

\usepackage[ruled,linesnumbered]{algorithm2e}
\usepackage{enumitem}
\usepackage{makecell}
\usepackage{hyperref}

\setcopyright{acmlicensed}
\copyrightyear{2026}
\acmYear{2026}
\setcopyright{cc}
\setcctype{by-nc-nd}
\acmConference[SenSys '26]{ACM/IEEE International Conference on Embedded Artificial Intelligence and Sensing Systems}{May 11--14, 2026}{Saint Malo, France}
\acmBooktitle{ACM/IEEE International Conference on Embedded Artificial Intelligence and Sensing Systems (SenSys '26), May 11--14, 2026, Saint Malo, France}
\acmDOI{10.1145/3774906.3800493}
\acmISBN{979-8-4007-2309-4/2026/05}

\begin{document}

\title{Optimizing Feature Extraction for On-device Model Inference with User Behavior
Sequences}

\author{Chen Gong}
\email{gongchen@sjtu.edu.cn}
\orcid{0000-0003-0333-6418}
\affiliation{
    \institution{Shanghai Jiao Tong University}
    \city{Shanghai}
    \country{China}
}

\author{Zhenzhe Zheng}
\authornote{Zhenzhe Zheng is the corresponding author.}
\email{zhengzhenzhe@sjtu.edu.cn}
\orcid{0000-0002-5094-5331}
\affiliation{
    \institution{Shanghai Jiao Tong University}
    \city{Shanghai}
    \country{China}
}

\author{Yiliu Chen}
\email{chenyiliu@bytedance.com}
\orcid{0009-0002-2775-3566}
\affiliation{
    \institution{ByteDance}
    \city{Hangzhou}
    \country{China}
}

\author{Sheng Wang}
\email{wangsheng.john@bytedance.com}
\orcid{0009-0009-5554-6207}
\affiliation{
    \institution{ByteDance}
    \city{Hangzhou}
    \country{China}
}

\author{Fan Wu}
\email{fwu@cs.sjtu.edu.cn}
\orcid{0000-0003-0965-9058}
\affiliation{
    \institution{Shanghai Jiao Tong University}
    \city{Shanghai}
    \country{China}
}

\author{Guihai Chen}
\email{gchen@cs.sjtu.edu.cn}
\orcid{0000-0002-6934-1685}
\affiliation{
    \institution{Shanghai Jiao Tong University}
    \city{Shanghai}
    \country{China}
}

\renewcommand{\shortauthors}{Gong et al.}

\begin{abstract}
    Machine learning models are widely integrated into modern mobile apps to analyze user behaviors and deliver personalized services. Ensuring low-latency on-device model execution is critical for maintaining high-quality user experiences.
    While prior research has primarily focused on accelerating model inference with given input features, we identify an overlooked bottleneck in real-world on-device model execution pipelines: extracting input features from raw application logs.
    In this work, we explore a new direction of feature extraction optimization by analyzing and eliminating redundant extraction operations across different model features and consecutive model inferences.
    We then introduce AutoFeature, an automated feature extraction engine designed to accelerate on-device feature extraction process without compromising model inference accuracy. AutoFeature comprises three core designs:  
    (1) graph abstraction to formulate the extraction workflows of different input features as one directed acyclic graph, 
    (2) graph optimization to identify and fuse redundant operation nodes across different features within the graph;
    (3) efficient caching to minimize operations on overlapping raw data between consecutive model inferences.
    We implement a system prototype of AutoFeature and integrate it into five industrial mobile services spanning search, video and e-commerce domains. Online evaluations show that AutoFeature reduces end-to-end on-device model execution latency by 1.33$\times$-3.93$\times$ during daytime and 1.43$\times$-4.53$\times$ at night.
\end{abstract}

\ccsdesc[500]{Human-centered computing~Mobile computing}
\ccsdesc[500]{Computing methodologies~Machine learning}

\keywords{On-Device Machine Learning; User Behavior Analysis; Model Inference Acceleration; Feature Extraction Optimization}
\maketitle

\input{Contents/1-Introduction.tex}
\input{Contents/2-Background.tex}
\input{Contents/3-Design.tex}
\input{Contents/4-Evaluation.tex}
\input{Contents/5-RelatedWork}
\input{Contents/6-Conclusion}


\bibliographystyle{ACM-Reference-Format}
\bibliography{main.bib}

\input{Contents/7-Appendix}

\end{document}

%% file: Contents/1-Introduction.tex
\section{Introduction}
With the rapid advancements of mobile devices, machine learning (ML) models are increasingly integrated into modern mobile apps to deliver personalized services~\cite{sarker2021mobile, liu2016lasagna, DBLP:conf/imc/AlmeidaLMDLL21, DBLP:conf/www/XuLLLLL19, gong2024ode, gong2023store}. 
Unlike traditional large-scale vision or language models that use \textit{static} input features (e.g., image pixels or tokens embeddings), real-world ML models deployed on mobile devices for industrial apps are typically smaller and rely on \textit{dynamic} input features extracted from evolving user behavior sequences to capture user intent (e.g., genre list of the last 5 watched videos reflect shifting interests).
Typical examples include customized product advertisements on e-commerce platforms~\cite{cheng2016wide, covington2016deep}, video recommendation and preloading in video apps~\cite{yeo2018neural, mehrabi2018edge, tran2018adaptive} and search result ranking in search engines~\cite{karmaker2017application, ziakis2019important, zhang2018towards}.
\begin{figure}
    \centering
    \includegraphics[width=1.\linewidth]{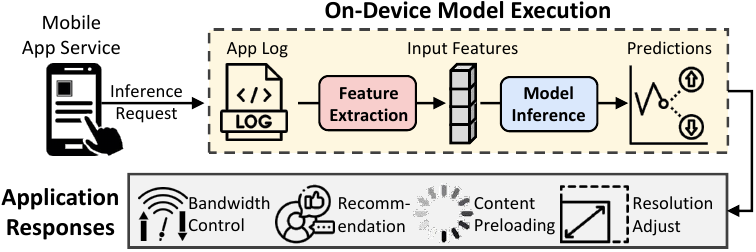}
    \vspace{-0.3cm}
    \caption{Workflow of real-world mobile application services.}
    \Description{Workflow of real-world mobile application services.}
    \vspace{-0.3cm}
    \label{fig: background}
\end{figure}

For on-device deployment, ensuring low-latency model execution is essential for maintaining both service quality and user experience. 
In real-world mobile service workflows as shown in Figure \ref{fig: background}, the device first extracts input features from historical user behaviors recorded in application logs (app log), and then performs model inference to predict user's current intent, with the ultimate goal of generating personalized app responses (e.g., preload suitable next-to-watch videos). 
High latency of on-device model execution not only blocks follow-up app responses at the system level, but also causes stuttering user experiences. Empirical studies show that even a slight 0.15\% increase in stuttering rate can lead to the loss of 9 million app users~\cite{u-apm}. 
Therefore, on-device ML model execution is expected to be imperceptible to users, ideally within 30 ms to match the human perception range of 30-60 FPS~\cite{deering1998limits}.

\textbf{Feature Extraction Bottleneck.}
For vision and language models, significant efforts have been made to optimize the efficiency of on-device model inference stage from both algorithm and hardware aspects~\cite{tang2023lut, guo2021mistify, kong2023convrelu++, jiang2020mnn, niu2021dnnfusion, khani2023recl, jeong2022band, wang2021asymo, yuan2022infi, li2020reducto, gong2025optimizing, gong2024delta}. 
However, our analysis of on-device model execution pipelines in real-world mobile apps reveals an overlooked bottleneck: \textit{extracting user features from raw app logs accounts for 61-86\% of the total model execution latency}.
This issue stems from three key factors~(elaborated in \S\ref{sec: feature extraction bottleneck}): 
(i) A large proportion of input features (74\% on average) required by on-device models for mobile apps are user features to reflect various user behaviors;
(ii) Extracting each user feature involves multiple resource-intensive operations on raw behavior data in app logs;
(iii) The on-device model inference is relatively fast due to model size limits and mature optimization techniques.

\textbf{Motivation.}
In this work, we delve into a crucial but unexplored direction,  \textit{feature extraction optimization}, to accelerate on-device model execution for mobile apps without sacrificing accuracy.
Our core insight is that many input features required by on-device models are extracted from overlapping user behavior data in app logs~(\S\ref{sec: optimization opportunities}).
This implies that redundant operations exist in extracting
(i) different input features within each model execution, and  (ii) identical input features across consecutive model executions.
These redundancies motivate us to alleviate feature extraction bottleneck by eliminating unnecessary data processing operations.

\textbf{Overall Design.}
We introduce AutoFeature, an efficient feature extraction engine for faster on-device model execution by eliminating redundant operations across different input features and consecutive model executions.
For an ML model deployed in the mobile app, AutoFeature represents its feature extraction process as a directed acyclic graph, termed \textit{FE-graph}, where source node denotes raw app log data, each target node denotes a feature and they are connected by a chain of operation nodes.
Next, AutoFeature optimizes the FE-graph from two aspects. 
(i) Inter-feature: within a single model execution, AutoFeature identifies and fuses redundant operation nodes across different features in the FE-graph;
(ii) Cross-execution: across consecutive inference requests, AutoFeature reuses intermediate results from previous model executions to eliminate redundant operations on overlapping data.
Since AutoFeature is designed to work independently before the model inference stage, it can be seamlessly integrated with any device operating systems, back-end mobile inference engines and ML models developed by different teams within a mobile app enterprise. It also complements previous efforts on model inference acceleration.

\textbf{Challenges and Our Solutions.}
AutoFeature addresses three major challenges in formulating and optimizing the FE-graph for single and multiple model executions.

\textit{First, constructing a unified FE-graph for automatic redundancy identification is non-trivial.}
Extracting input features from raw app logs is a complex process and existing literature lacks systematic analysis of it. Such opacity complicates the abstraction of feature extraction process into discrete, critical operation nodes that could facilitate automated redundancy detection across features.
To solve this, AutoFeature characterizes feature extraction as an information filtering process, which leverages multiple orthogonal conditions to progressively filter necessary information from raw app log data.  
Each feature's extraction process is then abstracted as a chain of operation nodes, each associated with different filtering conditions. In this way, any inter-feature redundancy can be systematically quantified by computing the intersections of filtering conditions of their operation nodes~(\S\ref{sec: graph generator}).

\textit{Second, efficiently eliminating redundancy across features is challenging.}
Intuitively, redundancies between any features with overlapping conditions can be eliminated by fusing their chains of operation nodes within the FE-graph. However, without careful designs on how to start and terminate the chain fusion process, the acceleration benefits can be offset by the introduced costs, including:
(i) operations on irrelevant app log data, as the fused filtering range of multiple conditions can be broader than their originally intended condition scope, and (ii) extra termination costs to separate the node outputs for fused features.
To address this, instead of treating each feature’s operation chain as a monolithic unit, AutoFeature decomposes it into multiple sub-chains with narrower condition per node, exposing finer-grained node fusion opportunities without condition scope expansion.
Further, a hierarchical filtering algorithm is proposed to progressively separate outputs for fused features based on their condition relations, reducing termination cost from $O\big(len(outputs)\!\times\!num(features)\big)$ to $O\big(len(outputs)\!+\!num(features)\big)$.

\textit{Third, minimizing redundant operations across consecutive model executions is not straightforward either.}
Although caching all intermediate results for each model execution could eliminate inter-execution redundancy, this approach is not always feasible and could cause potential app crashes due to the dynamic and limited memory space allocated to each ML model of each mobile app.
To optimally balance redundancy elimination and memory cost, AutoFeature formulates the caching decision as a classic knapsack packing problem, where the object is to maximize computational savings within a given memory budget.
A greedy policy is then proposed to prioritize intermediate results with the highest benefit-to-cost ratios, which can be efficiently measured in constant time complexity through our term decomposition technique~(\S\ref{sec: event evaluator}).

\indent \textbf{Contributions} of this work are summarized as follows:\\
$\bullet$ To the best of our knowledge, we are the first to unveil and analyze the feature extraction bottleneck in practical on-device ML model executions, exploring feature extraction optimization as a new research problem.\\
$\bullet$ We propose AutoFeature system to address feature extraction bottleneck without compromising model accuracy by eliminating redundant operations across different input features and consecutive model executions.\\
$\bullet$ We demonstrate AutoFeature's remarkable performance through extensive online evaluations in real-world mobile services, covering domains of search, video and e-commerce.

%% file: Contents/2-Background.tex
\section{Background and Motivation}
In this section, we first elaborate on-device model execution pipelines in mobile apps~(\S\ref{sec: model inference pipeline}). Then, we analyze feature extraction bottleneck in industrial mobile service workloads~(\S\ref{sec: feature extraction bottleneck}) and explore the optimization opportunities that motivate our work~(\S\ref{sec: optimization opportunities}).
\begin{figure*}
    \centering
    \includegraphics[width=0.9\linewidth]{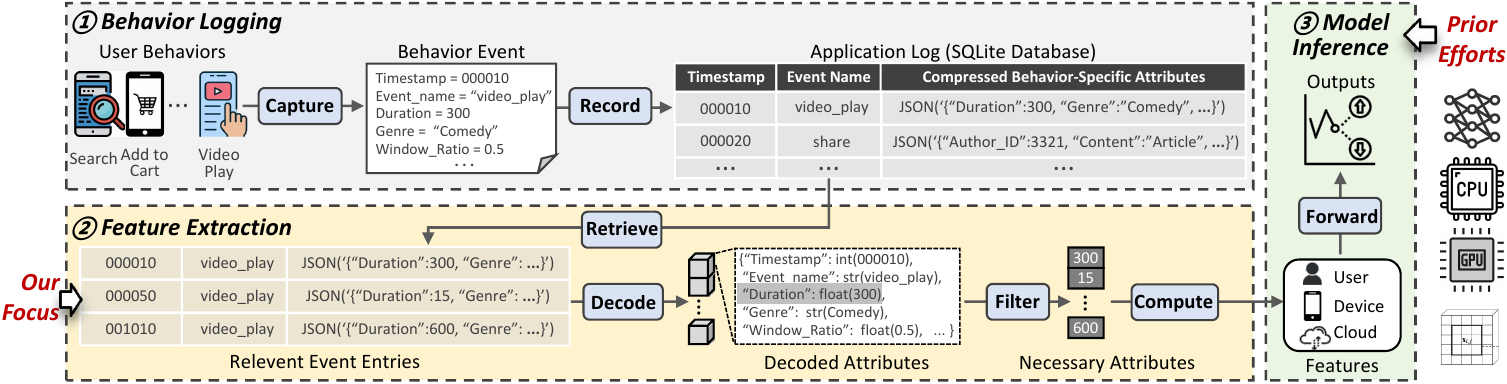}
    \vspace{-0.3cm}
    \caption{A complete on-device model execution pipeline in industrial mobile apps.}
    \Description{A complete on-device model execution pipeline in industrial mobile apps.}
    \label{fig: on-device model inference process}
    \vspace{-0.4cm}
\end{figure*}

\subsection{On-Device Model Execution Pipeline}
\label{sec: model inference pipeline}
As shown in Figure \ref{fig: on-device model inference process}, a complete on-device model execution pipeline in industrial mobile apps consists of three key stages that transform physical user behaviors into input features usable for ML models and ultimately into user intent predictions.

\textbf{Stage 1: Behavior Logging.} 
In mobile apps, each interaction behavior between user and the graphical user interface (GUI) can be captured as a behavior event (e.g., Search, Add-to-Cart, Video-Play). 
Each behavior event is a structured format of multiple attributes, including
\textit{behavior-independent} attributes (e.g., timestamp, event name) and
\textit{behavior-specific} attributes that vary by behavior type to provide detailed descriptions (e.g., duration and genre of Video-Play, item\_id and price of Add\_to\_Cart). 
Then, each behavior event is recorded as a single row in app log, which is typically managed using SQLite database~\cite{kreibich2010using, oh2015sqlite, gong2025optimizing} on both iOS and Android devices~\cite{feiler2015using, obradovic2019performance}. 
In Figure \ref{fig: attribute number CDF}, our analysis of 100 common behavior types from a popular video app shows that 50\% of user behavior types contain more than 25 attributes and 25\% contain over 85 attributes.
To manage this complexity and reduce storage costs, for each row of behavior event, behavior-independent attributes are stored in separate columns for data retrieval, while behavior-specific attributes are typically compressed into a single column\footnote{Storing behavior-specific attributes in separate columns would lead to excessive null values in app log and high storage cost~\cite{gong2025optimizing}, as different behaviors have heterogeneous attributes for description.}, as shown in the gray part of Figure \ref{fig: on-device model inference process}. 

\textbf{Stage 2: Feature Extraction.}
When an on-device model execution is invoked by a mobile service, the device extracts all input features required by the ML model to fully reflect the ongoing user context, including:
\textit{(i) user features} to summarize various user behavior types over different time periods (e.g., average duration of videos watched over the past hour or day),
\textit{(ii) device features} to describe the current device state  (e.g., volume level, battery percent), and
\textit{(iii) cloud features} to supplement information from service providers (e.g., embeddings of user\_id and service\_id).
While device and cloud features are either readily accessible or pre-fetched from cloud in advance, the specific values of user features are dynamic with time and require real-time extraction from the latest app logs.

\textbf{Stage 3: Model Inference.}
Once all required features are extracted, the on-device model performs inference to produce predictions for subsequent system responses. This process has been systematically supported by well-established mobile ML engines such as TensorFlow Lite~\cite{tflite}, MNN~\cite{jiang2020mnn} and ByteNN~\cite{bytenn}, which are optimized for efficient execution on mobile hardware. 

\subsection{Feature Extraction Bottleneck}
\label{sec: feature extraction bottleneck}
\begin{figure}
    \centering
    \begin{minipage}[b]{0.465\linewidth}
        \includegraphics[width=\linewidth]{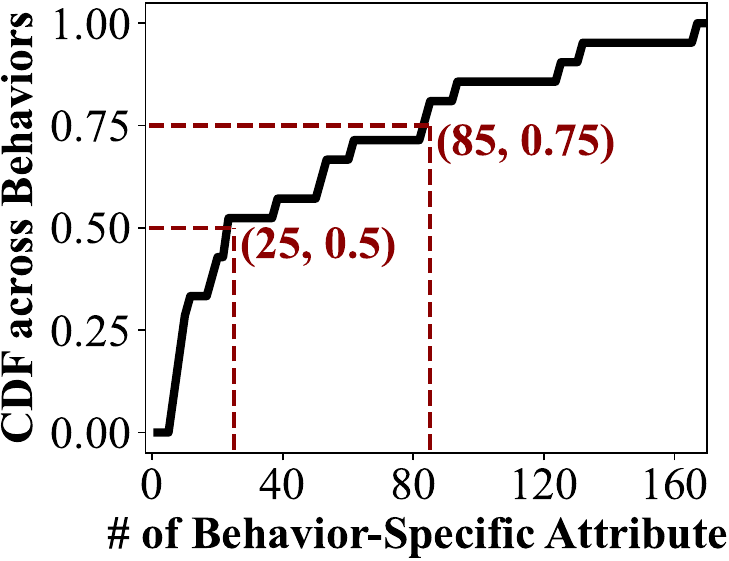}
        \vspace{-0.6cm}
        \caption{Attribute number of mobile user behaviors.}
        \Description{Attribute number of mobile user behaviors.}
        \vspace{-0.3cm}
        \label{fig: attribute number CDF}
    \end{minipage}
    \ \ 
    \begin{minipage}[b]{0.495\linewidth}
        \includegraphics[width=\linewidth]{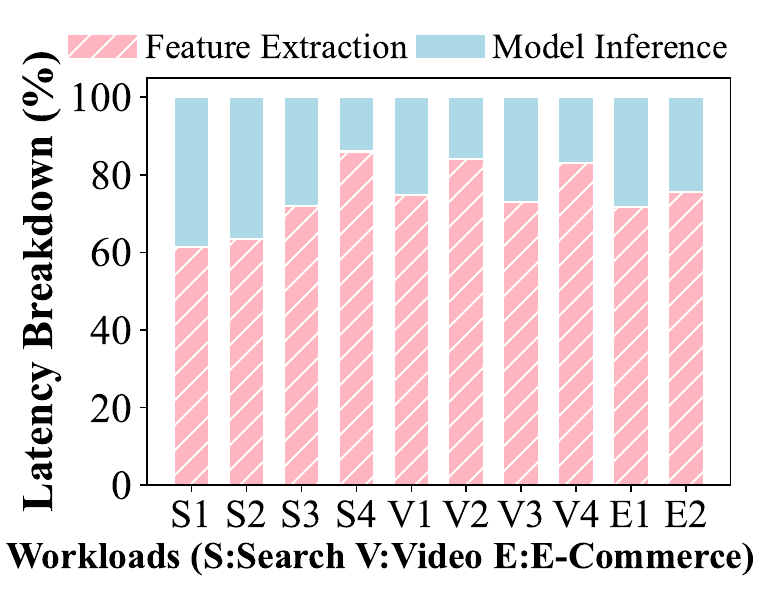}
        \vspace{-0.6cm}
        \caption{Time breakdown of on-device model execution.}
        \Description{Time breakdown of on-device model execution.}
        \vspace{-0.4cm}
        \label{fig: feature extraction bottleneck}
    \end{minipage}
\end{figure}
To achieve low-latency on-device model execution, prior academic research has predominantly focused on optimizing the model inference stage (see \S\ref{sec: related work}), as they targeted traditional vision and language models that use static input features like image pixels and word embeddings. 
In Figure \ref{fig: feature extraction bottleneck}, we observe real-world on-device model execution pipelines in industrial mobile apps across domains of search, video and e-commerce~(\S\ref{sec: methodology}) and break down their latencies. Our empirical study unveils an overlooked performance bottleneck: \textit{feature extraction alone accounts for $61$\%-$86$\% of the total end-to-end model execution latency}.
This significant bottleneck stems from three primary factors. 

\begin{figure}
    \centering
    \includegraphics[height=3cm]{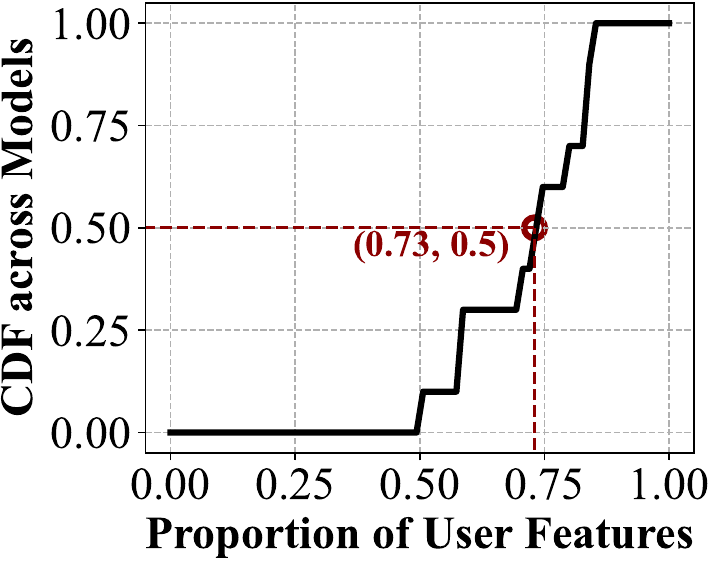}
    \ \ \ 
    \includegraphics[height=3cm]{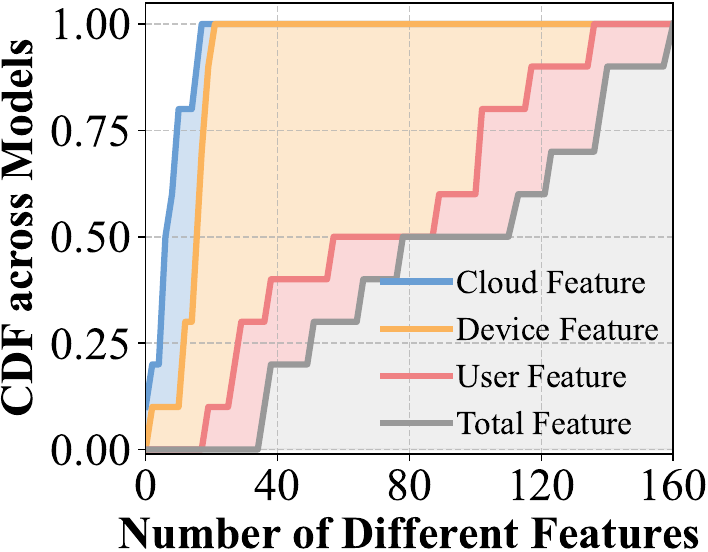}
    \vspace{-0.3cm}
    \caption{Proportion of user features (left) and numbers of various features (right) across on-device ML models.}
    \label{fig: high proportion of behavioral features}
    \vspace{-0.4cm}
\end{figure}
\begin{figure*}
    \vspace{-0.2cm}
    \subfigure[Features can require the same behavior type with overlapped time ranges.]{
        \includegraphics[height=3.2cm]{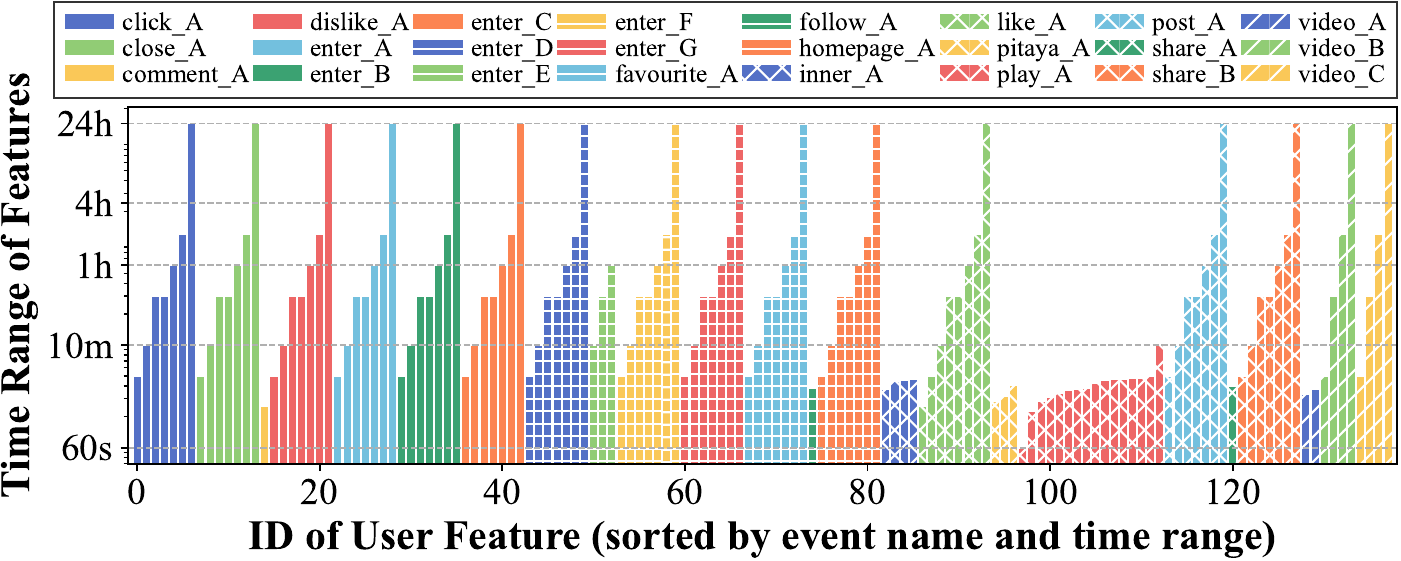}
        \label{fig: cross-feature redundancy}
    }
    \ \ 
    \subfigure[Cross-inference redundancy of different features (left) and models (right).]{
        \includegraphics[height=3.2cm]{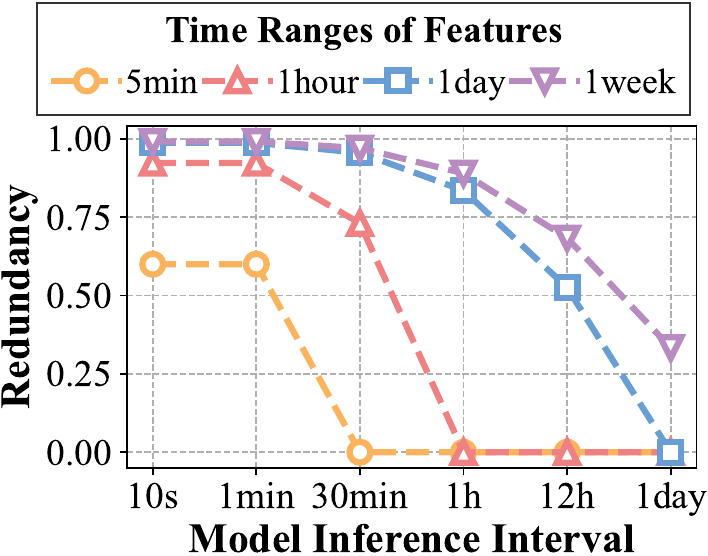}
        \ 
        \includegraphics[height=3.2cm]{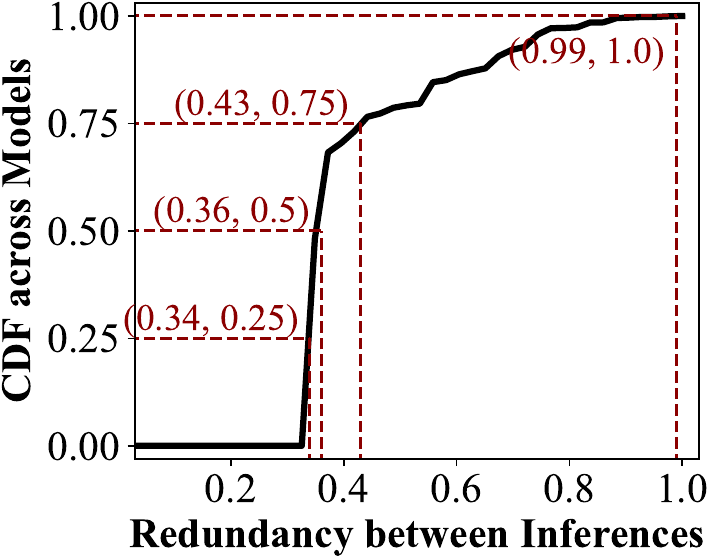}
        \label{fig: cross_inference_redundancy}
    }
    \vspace{-0.5cm}
    \caption{Analysis on real-world on-device ML models to reveal redundancy across features and inferences.}
    \Description{Analysis on real-world on-device ML models to reveal redundancy across features and inferences.}
    \vspace{-0.4cm}
\end{figure*}

\textbf{High Proportion of User Features.}
For popular mobile apps like TikTok and Taobao, the ML models practically deployed on devices are used to analyze user intent based on private and personal user behavior data. As a result, these models rely heavily on user features extracted from various behavior types across different time windows.
We analyze the proportions and numbers of user features across over 20 ML models deployed in mobile apps that collaborate with us and present the results in Figure \ref{fig: high proportion of behavioral features}. 
We notice that on average, user features constitute 73\% of the total input features required by ML models. Specifically, $50\%$ of on-device ML models require more than $60$ distinct user features and $20\%$ require as many as $110$ user features.

\textbf{Cumbersome User Feature Extraction.}
Extracting user features from raw app logs is resource-intensive and time-consuming, due to the misaligned granularity between data stored in app logs and required by model features.
As explained in \S\ref{sec: model inference pipeline} and visualized in Figure \ref{fig: on-device model inference process}, user behaviors are recorded by mobile apps at \textit{event level} in app logs, where each row corresponds to a specific behavior event and behavior-specific attributes are compressed in one column for efficient storage.  
However, each user feature typically correlates with only a few attributes of certain behavior types (i.e., \textit{attribute level}), such as average watching durations, genre list of videos watched over past hour. 
This is because different attributes are designed to describe the same behavior from diverse dimensions (e.g., watching duration, pause times, volume, genre, etc).
As a result, extracting each user feature requires multiple resource-intensive operations, including retrieving relevant event rows, decompressing detailed attributes and filtering necessary attributes for computation (elaborated in yellow part of Figure \ref{fig: on-device model inference process} and \S\ref{sec: graph generator}).

\textbf{Fast On-Device Model Inference.}
Compared with the high latency of feature extraction, the practical on-device model inference is relatively fast, typically completed within millisecond-level latency, which is due to three root causes:
\textit{(i) Model size limitation:} Mobile platforms have strict limits on app size (e.g., 2GB for iOS~\cite{ios_space} and 4GB for Android~\cite{android_space}), which limits the size of each ML model deployed by mobile app on native devices.
\textit{(ii) Simple task:} 
Most ML models practically offloaded to mobile devices handle lightweight but real-time prediction tasks for subsequent system responses, where compact models like decision trees~\cite{quinlan1996learning}, multi-layer perceptrons~\cite{ramchoun2016multilayer} and small neural networks~\cite{cheng2016wide} are qualified.
\textit{(iii) Advanced optimization:}
Years of research has led to highly efficient mobile inference engines~\cite{tflite, jiang2020mnn}, with the help of mature optimization techniques from hardware, operator, model and algorithm aspects~\cite{kong2023convrelu++, jiang2020mnn, niu2021dnnfusion, tang2023lut, guo2021mistify, khani2023recl, jeong2022band, wang2021asymo, gong2025optimizing, liu2025enabling}.

\subsection{Optimization Opportunities}
\label{sec: optimization opportunities}
In this work, we aim to address feature extraction bottleneck for real-world on-device model execution by eliminating redundant operations on overlapping data across both different input features and successive model inferences.

\textbf{Inter-Feature Redundancy.}
After scrutinizing ML models deployed in industrial mobile apps, we observe that many user features required by the same model can rely on overlapping behavior events, resulting in repeated operations on the same event rows in app log.
Specifically, in Figure \ref{fig: cross-feature redundancy}, we visualize the behavior types and time periods required by input features of a video recommendation model in TikTok, with feature names and behavior types anonymized for privacy. 
We observe that although the model requires 134 distinct user features, they are extracted from only 24 unique behavior types (indicated by colored bars and hatches).
Despite variations in their target time windows and behavior-specific attributes, raw event rows processed by different features can remain largely overlapped.
Such redundancy presents a substantial opportunity to optimize feature extraction process by fusing operations on overlapping events necessitated by different features.

\textbf{Cross-Inference Redundancy.}
Redundant operations also occur between consecutive model executions triggered by the same mobile service.
In mobile apps, user intent and preferences are dynamic, requiring frequent model inferences to deliver real-time, accurate system responses. When the interval between model executions is shorter than the time range required for feature extraction, event rows processed in previous execution remains relevant and reusable for next execution.
Figure~\ref{fig: cross_inference_redundancy} illustrates this pattern. For features based on the last 5 minutes of behavior, 60\% of relevant event rows can be reused when model inference is triggered every minute. This overlap increases to 90\% for features that rely on behavior within the last hour.
We further show that this redundancy is widespread by collecting cross-inference redundancies of ML models running online in mobile apps. As shown in Figure \ref{fig: cross_inference_redundancy}, 75\% on-device models exhibit over 34\% overlapping data between online inferences and 25\% exhibit overlap exceeding 43\%.

These findings highlight the inefficiencies in current feature extraction process, inspiring us to design a more efficient feature extraction engine for reducing on-device model execution latency without compromising model accuracy.

%% file: Contents/3-Design.tex
\section{AutoFeature Design}
In this work, we introduce AutoFeature, a universal feature extraction engine designed to automatically identify and eliminate redundant operations across different features and consecutive model executions, accelerating end-to-end execution of on-device ML models for mobile apps.

\subsection{Overview}
\label{sec: overview}

\begin{figure}
	\vspace{-0.3cm}
    \centering
    \includegraphics[width=0.9\linewidth]{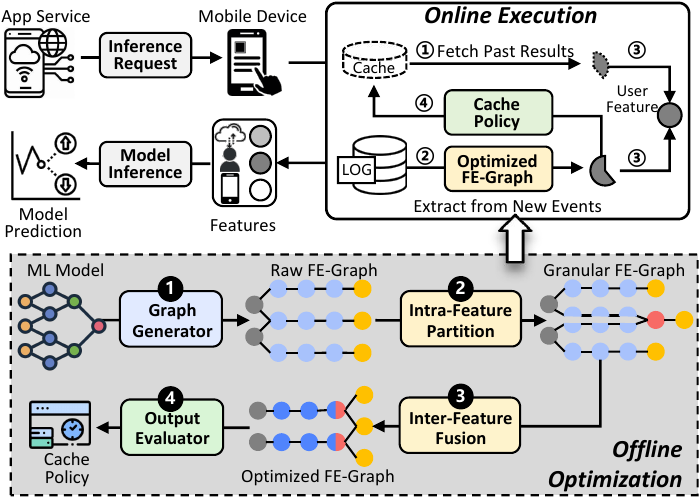}
    \vspace{-0.3cm}
    \caption{AutoFeature overview and workflow.}
    \Description{AutoFeature overview and workflow.}
    \vspace{-0.4cm}
    \label{fig: overall design}
\end{figure}

As depicted in Figure \ref{fig: overall design}, AutoFeature serves as an independent optimization layer that integrates seamlessly with existing on-device model execution pipelines. It complements existing optimizations on model inference stage and operates in two phases: offline optimization and online execution.

\textbf{Offline Optimization.}
When a new ML model is downloaded by mobile device via an app update, AutoFeature performs a one-time offline optimization to restructure the feature extraction workflow, which includes three main components.
First, {\ding{182}}\textit{graph generator} formulates the feature extraction workflow as a directed acyclic graph (FE-graph), where each source node of app log and each target node of an input feature is connected by a chain of critical operation nodes~(\S\ref{sec: graph generator}).
Second, AutoFeature optimizes the FE-graph to eliminate redundant operations across features~(\S\ref{sec: graph optimizer}), including 
{\ding{183}}\textit{intra-feature partition} to decompose each feature's operation chain into smaller sub-chains to expose finer-grained fusion opportunities for redundant operation nodes, and 
{\ding{184}}\textit{inter-feature fusion} to judiciously fuse operation nodes with overlapping inputs to eliminate redundancy and carefully design termination node to minimize output separation cost for fused node.
Third, {\ding{185}}\textit{output evaluator} valuates each node's outputs based on the ratio of potential computation savings to memory footprint. A greedy caching policy is then applied to select high-value intermediate results for reuse during online execution~(\S\ref{sec: event evaluator}).

\textbf{Online Execution.}
At run time, when an model execution request is issued, AutoFeature accelerates feature extraction through the following steps:
{\ding{172}} fetching previously computed intermediate results from the cache,
{\ding{173}} extracting missing intermediate results for newly logged user behavior data using the optimized FE-graph, 
{\ding{174}} assembling cached and newly computed results to reconstruct real-time user features, and
{\ding{175}} updating intermediate results in cache based on the cache policy and memory constraints.



\subsection{Graph Generator: Automated Redundancy Identification}
\label{sec: graph generator}
Given an on-device ML model, AutoFeature first needs to model the extraction process of each feature as a chain of discrete operation nodes, enabling systematic identification of redundancies across features.  
    Such design is inspired by successful modern ML frameworks (e.g., TensorFlow, PyTorch, MNN), which compile models into computational graphs to enable operator fusion and scheduling. Similarly, as a feature extraction engine, AutoFeature aims to abstract the entire feature extraction workflow into a graph for systematic redundancy identification and elimination.
To achieve this, we characterize the feature extraction process as a form of information filtering~\cite{belkin1992information}, which progressively removes unwanted information from an information stream using filtering conditions. 
Through an empirical analysis of user features required by on-device ML models, we conclude that any user feature can be defined by a set of four orthogonal conditions:
\begin{equation}
    <event\_names, time\_range, attr\_name, comp\_func>,
    \nonumber
\end{equation}
where \textit{event\_name} specifies the behavior types required by the feature, 
\textit{time\_range} defines the historical time window considered by the feature, 
\textit{attr\_name} denotes the specific attributes needed from such behaviors, and 
\textit{comp\_func} decides the computation function used to summarize behavior attributes.
\begin{figure}
    \centering
    \vspace{-0.3cm}
    \includegraphics[width=0.9\linewidth]{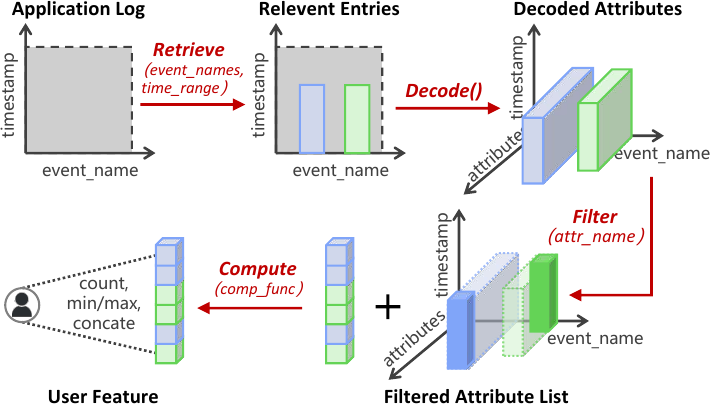}
    \vspace{-0.3cm}
    \caption{Four atomic operations of feature extraction.}
    \Description{Four atomic operations of feature extraction.}
    \vspace{-0.4cm}
    \label{fig: node abstraction of feature extraction}
\end{figure}

\textbf{Graph Formulation.}
The above definition allows us to divide each feature extraction process into four atomic operations, each corresponding to a specific condition to progressively extract necessary data from raw app logs. For better understanding, we provide a concrete example in Figure \ref{fig: on-device model inference process} and a high-level illustration in Figure \ref{fig: node abstraction of feature extraction}. The four atomic operations are as follows:

\noindent $\bullet$ \textit{Retrieve(event\_names, time\_range)}: 
Relevant behavior events required by the feature are retrieved from app logs to device memory according to the specified conditions of \textit{<event\_names, time\_range>}. 
This operation is typically implemented as a database query\footnote{``SELECT * FROM applog\_file WHERE event\_name IN \{\textit{event\_names}\} AND timestamp > \{(current\_time - \textit{time\_range})\}''} and data I/O between storage and memory is the primary overhead.

\noindent $\bullet$ \textit{Decode()}: 
For each retrieved row of behavior event, its detailed behavior-specific attributes are decoded from a compressed format. The decoding function is determined by the compression approach during behavior logging, typically implemented with lightweight data transformation tools like JSON parsing~\cite{pezoa2016foundations}. CPU dominates the overhead of this step.

\noindent $\bullet$ \textit{Filter(attr\_names)}: Next, necessary attributes are filtered from the decoded attributes and further converted into a computable format like C array or Python list.

\noindent$\bullet$  \textit{Compute(comp\_func)}: Finally, the filtered attributes are computed into the final input feature using the specified computing function. Common functions include count, average, concatenation to summarize user behaviors over a time period in different granularity.

\textbf{Redundancy Identification.}
In this way, the extraction workflows of multiple input features can be formulated as one directed acyclic graph. The source node corresponds to raw app log, each target node represents a user feature and they are sequentially connected by four critical operation nodes with distinct conditions.
Given an FE-graph, redundancy across any features can be identified by directly computing the set intersections of their conditions for each type of operation nodes.
We classify inter-feature redundancy into three levels based on the condition overlapping degree.
\textit{(i) No redundancy}: features have completely disjoint \textit{<event\_names, time\_range>} conditions, meaning no overlap in their relevant raw data (i.e., non-overlapping event rows in app log);
\textit{(ii) Partial redundancy}: features share intersected \textit{<event\_names, time\_range>} conditions, leading to redundancy in Retrieve and Decode operations;
\textit{(iii) Full redundancy}: features have identical \textit{<event\_names, time\_range>} conditions, implying duplicate Retrieve and Decode operation costs.
It is important to note that few features share totally same \textit{<event\_names, time\_range, attr\_name>} conditions as they depict identical behavior dimensions.

\subsection{Graph Optimizer: Inter-Feature Redundancy Elimination}
\label{sec: graph optimizer}
\begin{figure}
    \centering
    \vspace{-0.2cm}
    \includegraphics[width=0.9\linewidth]{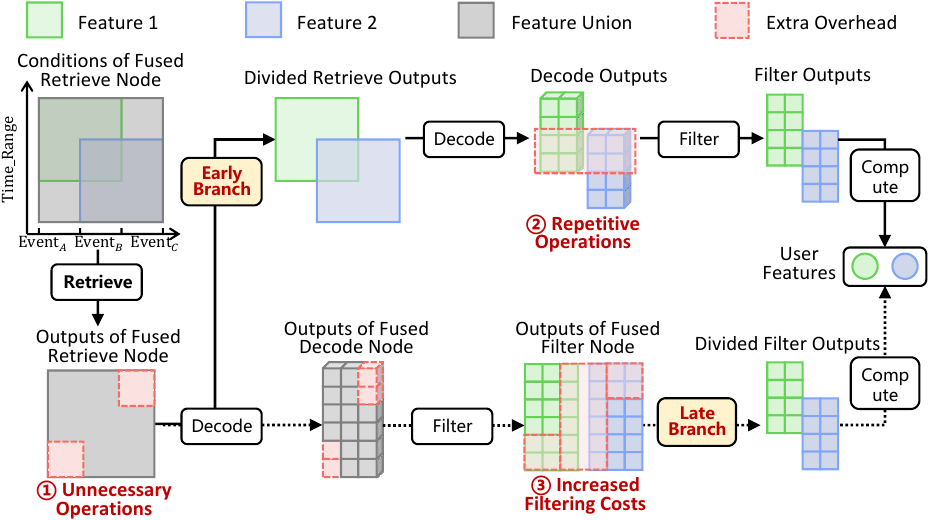}
    \vspace{-0.3cm}
    \caption{Additional cost introduced by feature fusion.}
    \Description{Additional cost introduced by feature fusion.}
    \vspace{-0.4cm}
    \label{fig: extra_cost_in_node_merging}
\end{figure}
	After quantifying redundancy across operations, a natural solution to eliminate such redundancy is fusing operation nodes with overlapping inputs for different features.
This involves merging their nodes of the same operation type into a single node, and setting  condition of such fused node as the union of the original conditions. 
    Despite conceptually simple, this method faces two primary challenges in how to start and terminate the chain fusion process to fully unleash the optimization potential.

\textit{Overgeneralized Conditions}: 
Chain fusion begins with fusing \textit{Retrieve} nodes, which have two orthogonal conditions: \textit{<event\_names>} and \textit{<time\_range>}. Thus, the set union of different \textit{Retrieve} nodes' conditions often results in a broader condition scope than the originally intended one. 
As illustrated in the left part of Figure \ref{fig: extra_cost_in_node_merging}, the condition union of feature 1 and 2 (gray area) is larger than the originally intended condition scope (green and blue areas), which incurs extra operations \ding{172} on irrelevant data (red area) for all subsequent operation nodes.

\textit{Optimal Termination Point}:
Each fused operation chain has to terminate with an extra \textit{Branch} node to separate outputs for different features before their \textit{Compute} operations, ensuring accurate feature extraction.
However, determining the optimal termination point is non-trivial. Early termination could leave redundancy in later operation nodes, while late termination causes excessive branching costs. 
As shown in Figure \ref{fig: extra_cost_in_node_merging}, early termination after \textit{Retrieve} node leaves repetitive operations \ding{173} on overlapping data, whereas delaying termination requires each feature to filter relevant data from more intermediate results \ding{174}.

To address these challenges, AutoFeature introduces a structured optimization strategy, which consists of two key steps: intra-feature chain partition and inter-feature chain fusion, to eliminate redundancy while maximizing computational efficiency.

\textbf{Intra-Feature Chain Partition.}
The root cause of unnecessary operations during chain fusion lies in the orthogonality of conditions of \textit{Retrieve} nodes (i.e., \textit{event\_names} and \textit{time\_range}). Therefore, AutoFeature proposes to decompose each \textit{Retrieve} node into multiple sub-nodes, where each sub-node retains the original \textit{time\_range} condition but is assigned a distinct \textit{event\_name} condition.
This decomposition partitions each feature extraction chain into multiple sub-chains with finer-grained conditions per node, allowing more precise chain fusion. By ensuring that only sub-chains with identical \textit{event\_name} are fused, AutoFeature prevents irrelevant data from entering the pipeline.
\begin{figure}
    \centering
    \vspace{-0.2cm}
    \includegraphics[width=0.44\linewidth]{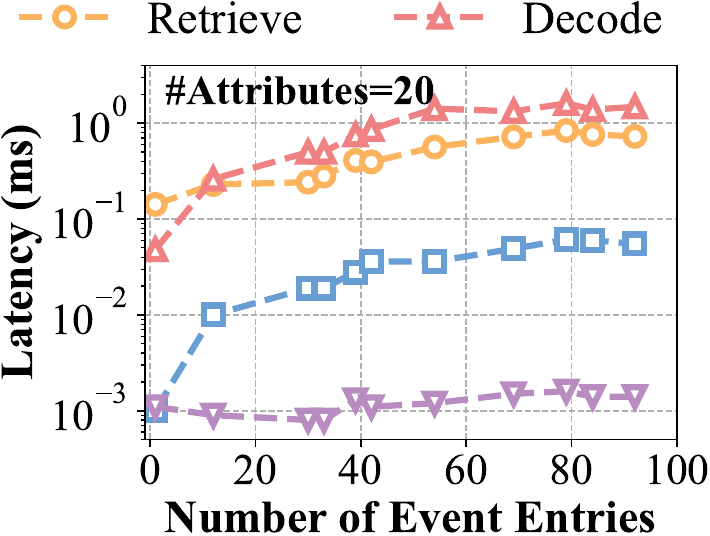}
    \ \ 
    \includegraphics[width=0.44\linewidth]{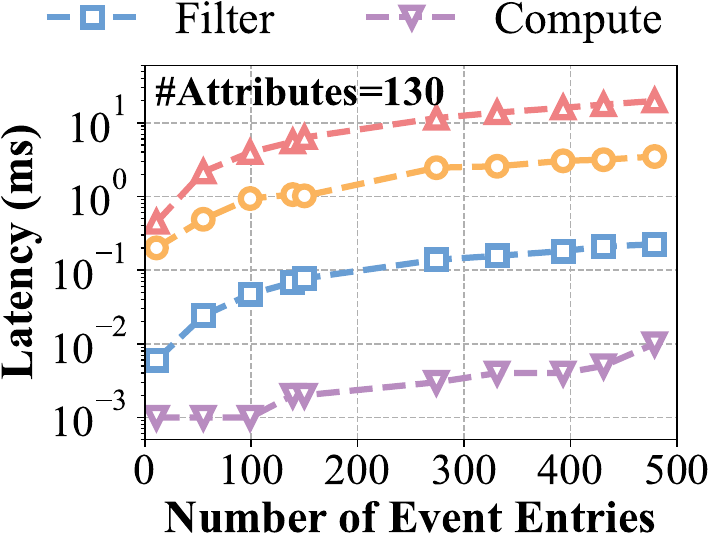}
    \vspace{-0.3cm}
    \caption{Latency breakdown of extracting user features from behavior events with different attributes.}
     \Description{Latency breakdown of extracting user features from behavior events with different attributes.}
    \vspace{-0.4cm}
    \label{fig: per-operation time}
\end{figure}

\textbf{Inter-Feature Chain Fusion.}
Further, we introduce ``branch postposition'' concept to maximize redundancy elimination and a hierarchical filtering algorithm to minimize termination overhead.

\noindent \textit{$\bullet$ Branch Postposition Concept}.
A key observation from our system evaluation is that \textit{Retrieve} and \textit{Decode} nodes dominate feature extraction time. As shown in Figure \ref{fig: per-operation time}, these nodes consume 15$\times$ more time than \textit{Filter} nodes and 300$\times$ more time than \textit{Compute} nodes. 
To fully eliminating the redundancy of computationally expensive nodes, AutoFeature delays the insertion of \textit{Branch} nodes until just before the \textit{Compute} node.

\begin{figure}
    \centering
    \includegraphics[width=0.9\linewidth]{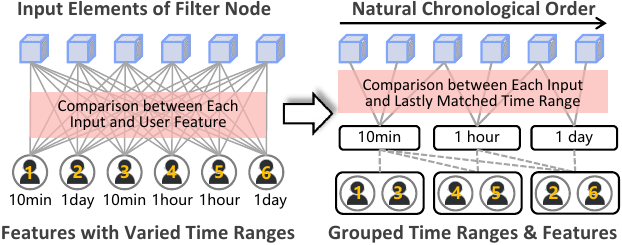}
    \vspace{-0.3cm}
    \caption{Comparison between original design and hierarchical filtering. Each line denotes a comparison.}
    \Description{Comparison between original design and hierarchical filtering. Each line denotes a comparison.}
    \vspace{-0.4cm}
    \label{fig: hierarchical algorithm}
\end{figure}

\noindent $\bullet$ \textit{Hierarchical Filtering Algorithm.}
Instead of appending separate \textit{Branch} nodes for each feature, AutoFeature proposes to integrate them into the fused \textit{Filter} node. As a result, for each input element, the \textit{Filter} node checks whether it satisfies the condition of each fused feature and extracts necessary attributes for different fused features. 
However, such direct integration increases the \textit{Filter} node's computational complexity to $O\big(len(input)\times num(feature)\big)$, which is computationally expensive when facing numerous features or massive relevant behavior events as shown in Figure \ref{fig: hierarchical algorithm}. 

To reduce this overhead, AutoFeature employs a hierarchical filtering algorithm based on two key observations: 
(i) Behavior events are stored in chronological order in app log, implicating that the outputs of each operation node are also time-ordered;
(ii) Model features typically consider meaningful, periodic time ranges (e.g., the past 1 hour, 1 day, 1 week), leading to grouped \textit{time\_range} conditions. 
Using these properties, AutoFeature could pre-compute a reverse mapping from each time range to the corresponding features and their required attributes, which can be pre-constructed offline.
At runtime, the fused \textit{Filter} node hierarchically filters attributes from input elements for each feature through the following steps, as visualized in Figure \ref{fig: hierarchical algorithm}:
\begin{itemize}[topsep=0cm, leftmargin=0.2cm]
    \item Identify the closest matching time range for each input element based on its timestamp attribute and pre-computed time ranges (i.e., keys of reverse mapping).
    \item Extract necessary attributes for each feature (i.e., values of reverse mapping) associated with the matched time range.
\end{itemize}
Since input elements arrives in chronological order, AutoFeature can begin comparison from the matching time range of previous input element, reducing the overall complexity to $O\big(len(inputs) + len(set(time\_ranges)\big)$. This achieves a speedup proportional to the number of fused features even when their time ranges differ.

\subsection{Event Evaluator: Inter-Inference Redundancy Minimization}
\label{sec: event evaluator}
In industrial deployment, mobile services require frequent on-device model executions to keep up with the latest user intention and ensure high-quality service.
Successive model executions often involve extracting user features from overlapping behavior events.
    To eliminate such temporal redundancy, the classic solution is to cache the valuable intermediate results during feature extraction for future reuse,
(i.e., attributes decoded and extracted by each feature. However, this solution faces challenges in real-world mobile apps). 
Mobile devices typically support multi-app execution, and the memory allocated to one ML model of a single mobile app can be limited and dynamic, making caching all intermediate results not always feasible. 

To find the sweet spot between memory footprint and redundancy elimination, AutoFeature formulates the determination of which intermediate results to cache as a classic knapsack problem. It further incorporates a greedy cache policy to provide theoretically guaranteed performance under various memory budgets.

\textbf{Caching Content Valuation.}
We first discuss which type of intermediate results should be cached. As analyzed previously, \textit{Decode} and \textit{Retrieve} operations dominate the feature extraction cost, which suggests that caching should prioritize these operation nodes' outputs for maximal computation savings.
Therefore, AutoFeature caches at behavior level, i.e., selecting certain behavior types and caching all their events' necessary attributes, rather than at feature level, i.e., caching attributes for certain features.
This approach eliminates the need to re-execute \textit{Decode} and \textit{Retrieve} nodes on the same behavior events to extract those uncached attributes.

Next, we define two metrics for each behavior type $E_i$ to quantify its caching utility and cost for further problem formulation:

\noindent $\bullet$ Utility $U(E_i)$ is quantified by the computational savings achieved by caching all necessary attributes of the retrieved events of $E_i$, primarily coming from redundant operations on overlapping events between consecutive inferences:
\begin{equation}
        U(E_i)=Num\_Overlap(E_i) \times Cost\_Opt(E_i),
    \nonumber
\end{equation}
where $Num\_Overlap$ denotes the number of overlapped events between consecutive executions and $Cost\_Opt$ denotes the operation cost per event.

\noindent $\bullet$ Cost $C(E_i)$ is the memory space required to cache attributes of events that are processed in current model execution:
\begin{equation}
        C(E_i)=Num(E_i) \times Size(E_i),
    \nonumber
\end{equation}
where $Num(E_i)$ denotes the number of events processed in current execution and $Size(E_i)$ denotes the size per event.

\textbf{Problem Formulation.}
Given a set of behavior types $\{E_i\}_{i=1}^N$ and a memory budget $M$, we aim to derive the optimal cache policy $\mathcal{P}^*$ to maximize redundancy reduction while staying within memory constraints:
\begin{equation}
        \mathcal{P}^*\!=\!\mathop{\arg\max}_{\mathcal{P}_i\in\{0, 1\}} \sum_{i=1}^{N}\big[\mathcal{P}_i\times U(E_i)\big]\ \ 
        \mathrm{s.t.,} \sum_{i=1}^N\big[\mathcal{P}_i\times C(E_i)\big]\!\le\!M,
    \label{eq: cache problem}
\end{equation}
where $\mathcal{P}_i\!\in\!\{0, 1\}$ is a decision variable indicating whether to cache the behavior type $E_i$. 
Essentially, the optimization problem (\ref{eq: cache problem}) is a classic knapsack problem~\cite{martello1987algorithms, salkin1975knapsack} that can be solved by dynamic programming algorithm with $O(NM)$ complexity~\cite{pferschy1999dynamic}. However, both of the memory constraint $M$ and the number of overlapped events between model executions are dynamic. They require solving the problem for each real-time model execution and is thus impractical for industrial deployment.

\textbf{Greedy Policy.}
AutoFeature incorporates a greedy cache policy that prioritizes behavior types based on their utility-to-cost ratio. Specifically, during each feature extraction process, AutoFeature sorts different behavior types by $U(E_i)/C(E_i)$ in a descending order, and iteratively caches attributes for behavior events with the highest ratio until the memory budget is exhausted. 
The above greedy policy provides both performance and efficiency guarantees.
Theoretically, previous research has proved that greedy solutions can achieve a 2-approximation ratio for the knapsack packing problems~\cite{chekuri2005polynomial}, which ensures a robust performance across various memory limitations imposed to each ML model. 
Further, the dynamic ratio $U(E_i)/C(E_i)$ can be computed with constant complexity in practice through term decomposition: 
\begin{equation}
        \begin{aligned}
        \frac{U(E_i)}{C(E_i)} \overset{(a)}{=} & \frac{Time\_Overlap(E_i)\times Freq(E_i)\times Cost\_Opt(E_i)}{Time\_Range(E_i)\times Freq(E_i)\times Size(E_i)}\\
        =& \underbrace{\frac{Time\_Overlap(E_i)}{Time\_Range(E_i)}}_\text{ Dynamic Term 1}\times \underbrace{\frac{Cost\_Opt(E_i)}{Size(E_i)}}_\text{Static Term 2},
        \end{aligned}
    \nonumber
    \label{eq: greedy cache policy}
\end{equation}
where Equation (a) represents the number of events as the multiplication of time range and the behavior occurrence frequency.
As a result, term 1 is dynamically determined by on-device model inference frequency and term 2 is static and can be recorded once in an offline manner. 

\textbf{Online Execution.}
To integrate this caching policy into online feature extraction, AutoFeature deploys an event evaluator that dynamically adjusts caching decisions at runtime. The feature extraction workflow follows these steps:
(i) Retrieve attributes of relevant behavior events from cache and update the feature's \textit{time\_range} conditions based on the timestamp of cached events;
(ii) Execute \textit{Retrieve} and \textit{Decode} operation nodes for missing attributes of newly logged events,
(iii) Merge cached and newly extracted attributes and execute \textit{Filter} and \textit{Compute} nodes to generate final features, 
(iv) Greedily cache events based on their up-to-date ratios of utility and cost, which are also leveraged to update cache when facing dynamic memory budget.


%% file: Contents/4-Evaluation.tex
\section{Evaluation}

\subsection{Experiment Setup}
\label{sec: methodology}

\textbf{Implementation.}
We have implemented a system prototype of AutoFeature and integrated it into the SDK of industrial mobile apps for evaluation\footnote{Updates to the model configurations like input features and model parameters are typically handled by downloading a new mobile app SDK through standard app updates. For this case, AutoFeature simply treats the updated configuration as a new model and re-runs its offline optimization phase within millisecond-level latency.}. AutoFeature is designed to operate without modifying the model inference process, ensuring that model trainers, inference engine developers and mobile users do not need to make any adjustments. 
In compliance with enterprise data privacy requirements, our evaluation primarily uses ByteNN~\cite{bytenn} as backend engine, which delivers comparable or superior performance to the open-source MNN~\cite{jiang2020mnn} through app-specific optimizations.

\begin{figure*}
    \centering
    \vspace{-0.4cm}
    \subfigure[User feature statistics: Each bar denotes a feature and each color represents a behavior type.]{
        \includegraphics[width=0.68\linewidth]{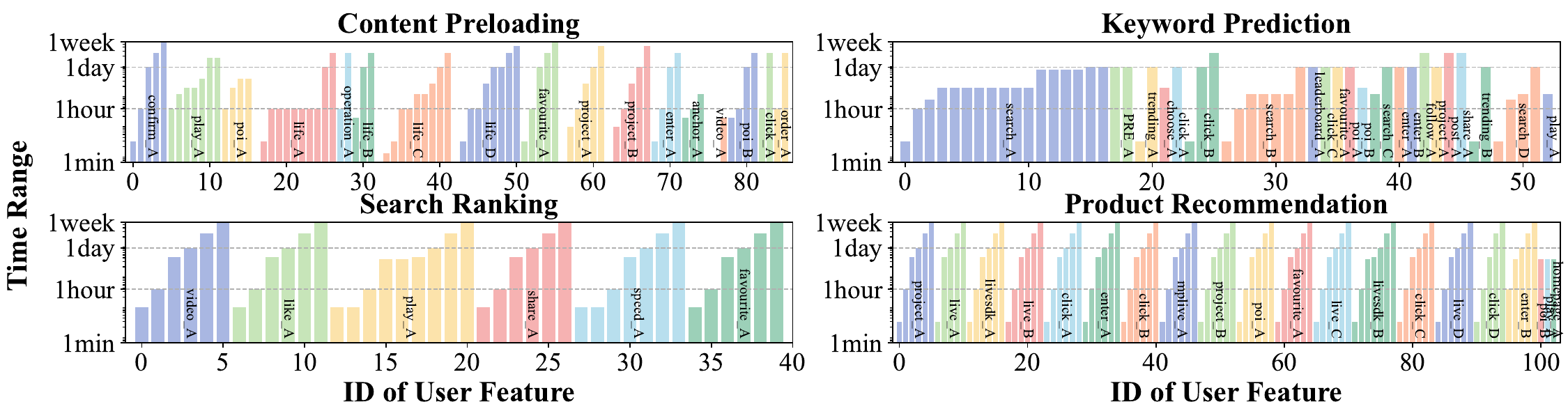}
        \label{fig: feature distribution}
    }
    \ \ \ \ \ 
    \subfigure[Inference frequency distribution.]{
        \includegraphics[width=0.20\linewidth]{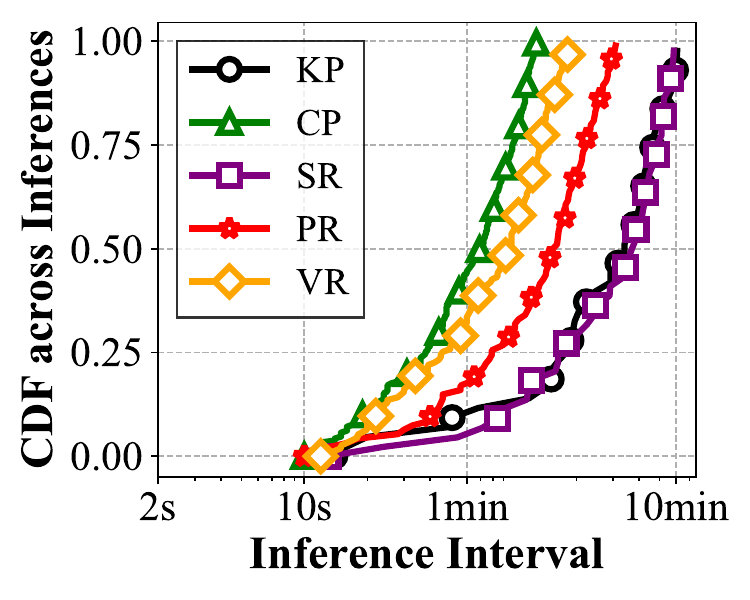}
        \label{fig: invocation frequency}
    }
    \vspace{-0.5cm}
    \caption{User feature definitions and online execution frequencies of on-device ML models in our evaluation.}
    \Description{User feature definitions and online execution frequencies of on-device ML models in our evaluation.}
    \vspace{-0.4cm}
    \label{fig: high-level description}
\end{figure*}

\textbf{Mobile Services.}
    Our evaluation was conducted on popular mobile apps from ByteDance, a company with billions of users, advanced on-device AI infrastructure and diverse mobile services.
To demonstrate the generality of AutoFeature, we evaluate its performance across five representative mobile services, spanning three typical domains of mobile apps: search engines, video apps and e-commerce platforms. 
    For video domain, TikTok (Douyin) is one of the world's most popular short-form video platforms~\cite{douyin_wiki}. For search domain, Toutiao~\cite{toutiao_wiki} is one of the most popular content discovery and searching platforms in China. For, e-commerce domain, Douyin is also a popular e-commerce platform.
    The tested services vary widely in scenarios, tasks, input user features and model execution frequencies, which are visualized in Figure \ref{fig: high-level description}.

\noindent $\bullet$ \textit{Content Preloading (CP)}:
Common in video apps like TikTok, this service decides which segment of video and comment to preload for ensuring seamless video watching experiences. The model relies on 86 user features derived from 27 distinct behavior types (e.g., video click, playback duration, sharing). 

\noindent $\bullet$ \textit{Keyword Prediction (KP)}:
Deployed in mobile search engines like Google, this service predicts likely search query keywords based on past query behavior and currently browsing content. It uses 53 user features to track 22 types of search-related behaviors (e.g., past search terms, click-through history, time spent on search results.).

\noindent $\bullet$ \textit{Search Ranking (SR)}:
This service ranks the returned search results to match current user preferences and improve searching relevance. It uses 40 user features to tracks 10 user behavior types related to searches such as interactions with ranked elements and engagement with search results.

\noindent $\bullet$ \textit{Product Recommendation (PR)}:
Widely used in e-commerce platforms like Taobao, this service offers personalized product recommendations and targeted advertisements. Its model extracts 103 user features across 21 types of commercial behaviors, including product browsing duration, cart additions, purchases, and search within product categories.

\noindent $\bullet$ \textit{Video Recommendation (VR)}: 
generates personalized video suggestions based on a user's viewing history and content preferences. The relevant user features and behavior types for this model were previously illustrated in Figure \ref{fig: cross-feature redundancy}.\\
As (i) modern mobile operating systems aggressively prioritize the currently active foreground app and (ii) on-device model inferences are mostly triggered when the user is interacting with the app, our testing app is guaranteed to be in the foreground and receive the dominant share of available hardware resources, limiting the interference from other co-running background apps. 

\begin{figure}
    \centering
    \includegraphics[width=0.8\linewidth]{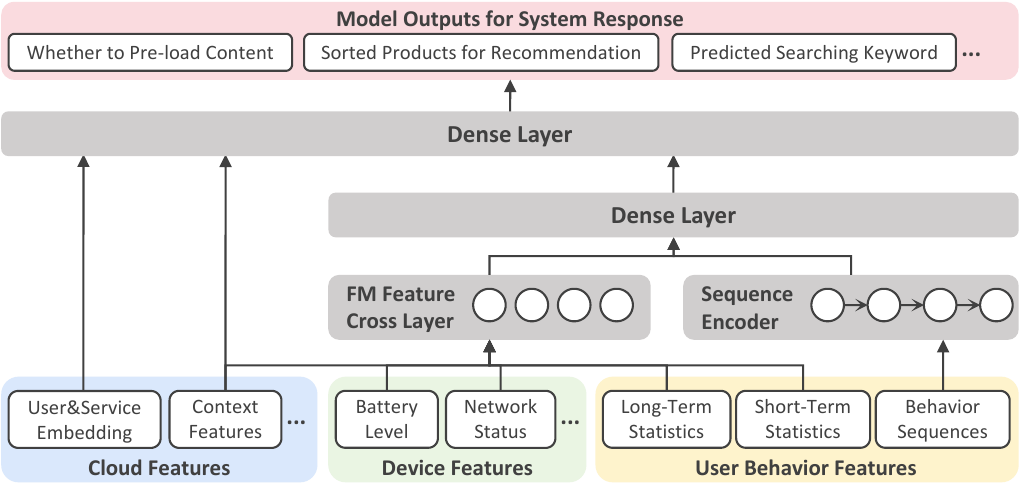}
    \vspace{-0.3cm}
    \caption{
        Model structure of common on-device ML models.
    }
    \Description{Model structure of common on-device ML models.}
    \vspace{-0.4cm}
    \label{fig: model architecture}
\end{figure}
\textbf{Model Architecture.}
    While we cannot disclose specific model architectures due to confidential requirement, we present a general structure of on-device models used by popular mobile services in Figure \ref{fig: model architecture}, which composes of three layers: \\
    $\bullet$ \textit{Input Layer}: As analyzed in \S\ref{sec: model inference pipeline}, An on-device model takes three categories of features as inputs: cloud features to provide global information, device features to describe the current device state and user features to summarize various historical user behaviors. \\
    $\bullet$ \textit{Processing Layer}: These features are then processed by different model layers. Statistical features of user behaviors and device features are passed into an factorization machine layer for feature crossing, while sequential behavior features are sent to a sequence encoder to capture temporal dynamics and periodical patterns.\\
    $\bullet$ \textit{Output Layer}: Finally, the combined feature outputs are passed through several dense layers to generate final predictions for personalized system responses.
\begin{figure}
    \centering
    \includegraphics[width=\linewidth]{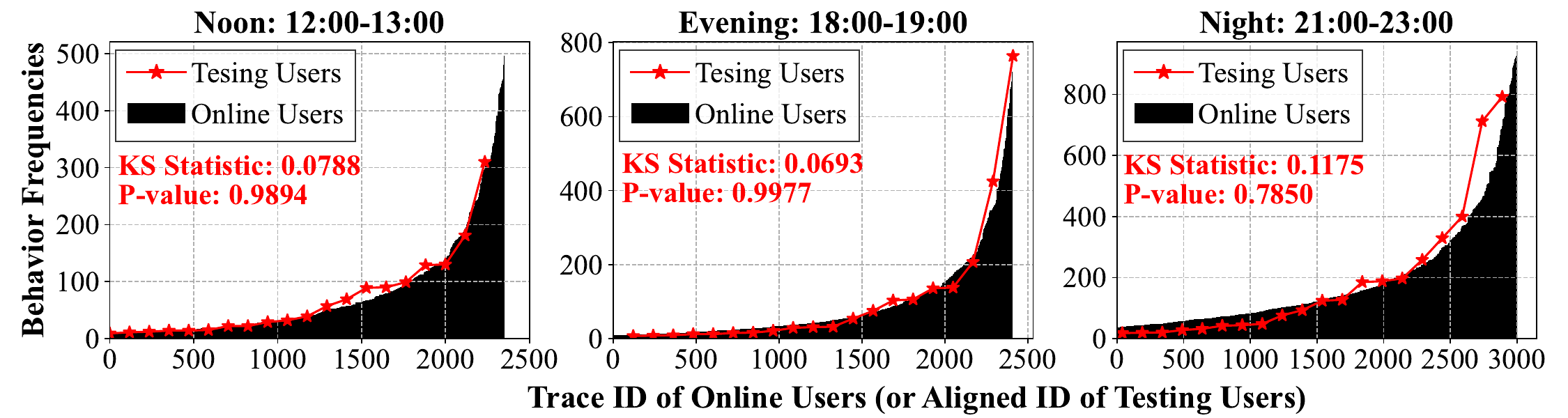}
    \vspace{-0.4cm}
    \caption{
        Statistically similar usage patterns between testing users and thousands of real-world users.
    }
    \Description{Statistically similar usage patterns between testing users and thousands of real-world users.}
    \vspace{-0.3cm}
    \label{fig: similar usage patterns}
\end{figure}
\begin{figure}
    \centering
    \includegraphics[width=0.5\linewidth]{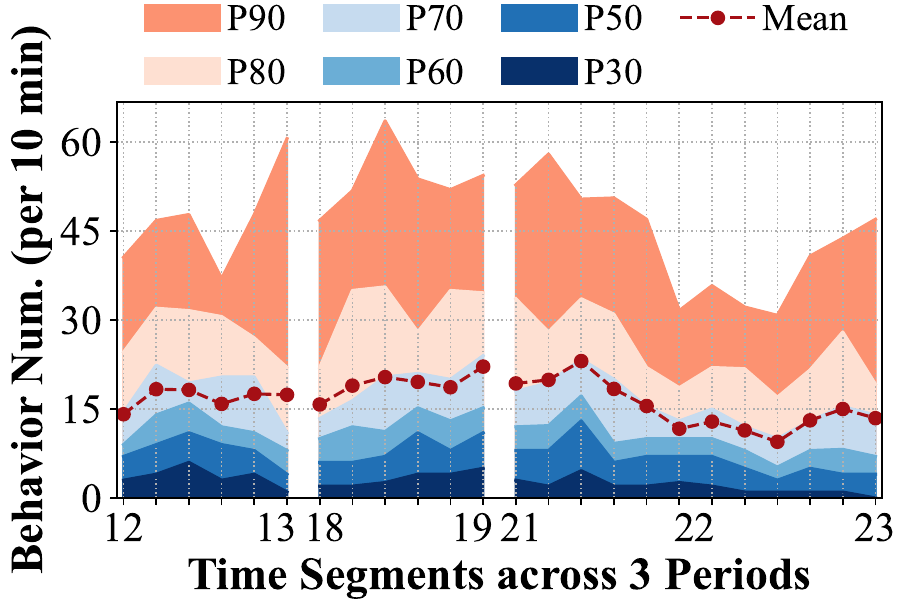}
    \vspace{-0.3cm}
    \caption{
        Specific behavior traces of testing users.
    }
    \Description{Specific behavior traces of testing users.}
       \vspace{-0.5cm}
    \label{fig: traces of all behaviors}
\end{figure}

\textbf{Testing Users.}
We evaluate AutoFeature's performance through online evaluation within real-world mobile apps. For each mobile service, we collect the end-to-end latency of on-device model execution of 10 testing users during their daily usage of the app across 2 days,
containing three common time periods: noon (12:00-13:00), evening (18:00-19:00) and night (21:00-23:00). 
    The limited test group stems from a necessary trade-off between operational cost and data representativeness. A fair and reliable comparison requires repeatedly running different feature extraction methods for every single real-time inference request. While this ensures identical user data and device state across methods, it significantly degrades service responsiveness and user experience. Limiting testing to 10 anonymous real-world users mitigates operational cost and financial loss, while still ensuring representativeness of the general population: \\
    $\bullet$ \textit{Statistically Similar Usage Patterns}: 
    We quantitatively validate that the test group's statistical usage patterns closely match those of massive real-world active users. Figure \ref{fig: similar usage patterns} visualizes the behavior frequencies of thousands of real-world user base (black bars) against 10 test users' 20 traces (red stars).
    We used the Kolmogorov-Smirnov (KS) test to compare these distributions. For different time periods, the KS statistic is extremely low ($0.079$-$0.118$) and the p-values ($0.785$-$0.998$) are significantly greater than the standard $0.05$ threshold, confirming that there is no statistically significant difference between the usage patterns of test users and overall user base.\\
    $\bullet$ \textit{Diverse Behavior Traces}: 
    Beyond statistical alignment, our 10 test users were selected to cover a wide range of activity intensity, demonstrating the generality of our evaluation.
    Figure \ref{fig: traces of all behaviors} illustrates the fine-grained behavior frequencies of test users, segmented into 10-minute intervals across different time periods.
	The top 10\% most active users (P90 traces) generate over 45 specific behaviors every 10 minutes, while the bottom 30\% of users (P30 traces) generate fewer than 5 behaviors per 10 minutes. Further details are 
    provided in Appendix \ref{sec: appendix}.
\\
\textbf{It is important to note that all data collection and measurement processes are conducted with the consent of testing users, ensuring full compliance with privacy regulations in both research and industry}.

\textbf{Baselines.}
To the best of our knowledge, AutoFeature is the first work to optimize feature extraction for on-device model execution of real-world mobile apps. Thus, we consider the following baselines and ablated versions of AutoFeature:
(i) \textit{w/o AutoFeature}: the industry-standard on-device feature extraction process, where each user feature is extracted independently without optimization, and 
(ii) \textit{w/ Fusion}: only the graph optimizer of AutoFeature is employed to eliminate inter-feature redundant operations,
(iii) \textit{w/ Cache}: only the cache policy of AutoFeature is applied to minimize inter-inference redundancy.
\begin{figure*}
    \centering
    \vspace{-0.4cm}
    \begin{minipage}[b]{0.80\linewidth}
        \subfigure[End-to-end latency of on-device model execution of different methods. We annotate the average speedups achieved by AutoFeature.]{
            \includegraphics[width=\linewidth]{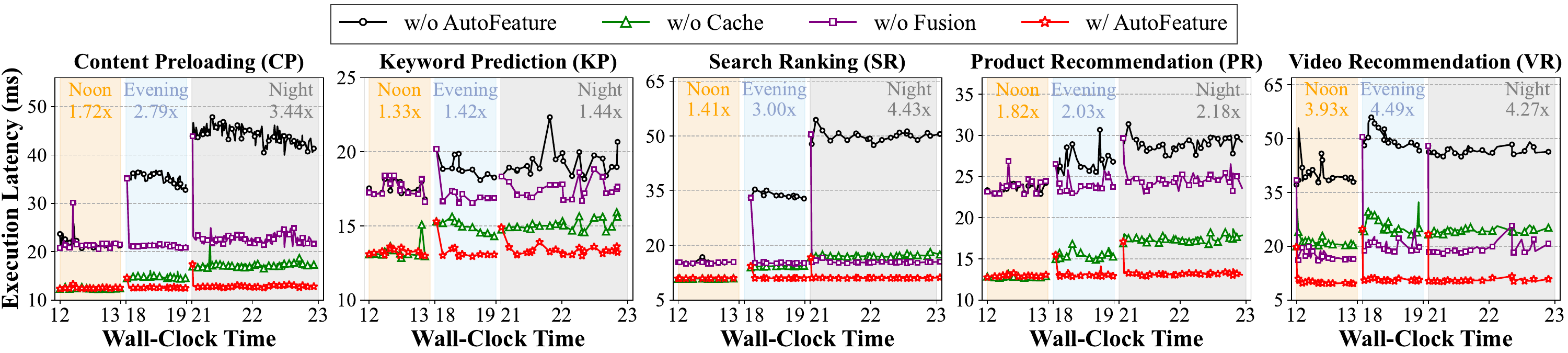}
        }
        \subfigure[Speedups of on-device model execution of different methods compared with industry-standard baseline (w/o AutoFeature).]{
            \includegraphics[width=\linewidth]{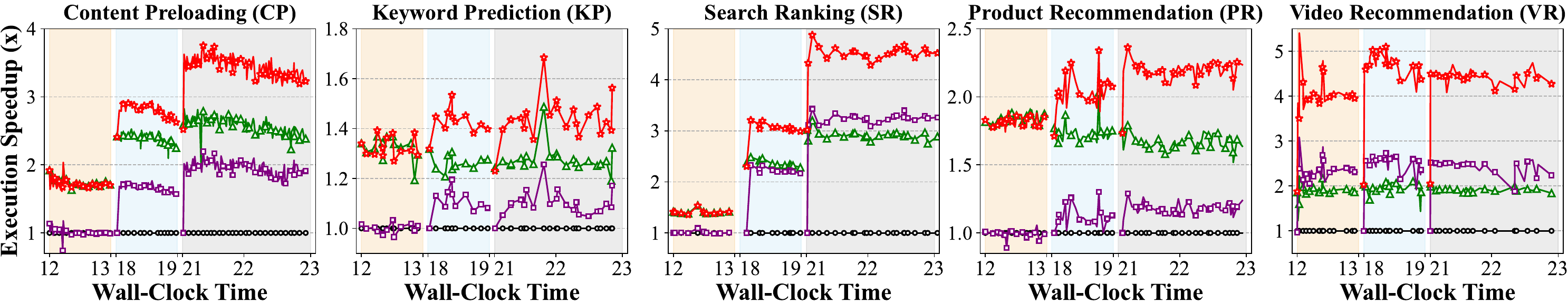}
        }
        \vspace{-0.3cm}
        \caption{Overall performance of different methods across varied time ranges and mobile services.}
        \vspace{-0.3cm}
        \label{fig: overall performance}
    \end{minipage}
    \ \ 
    \begin{minipage}[b]{0.184\linewidth}
        \subfigure[Offline: time consumption.]{
            \includegraphics[width=\linewidth]{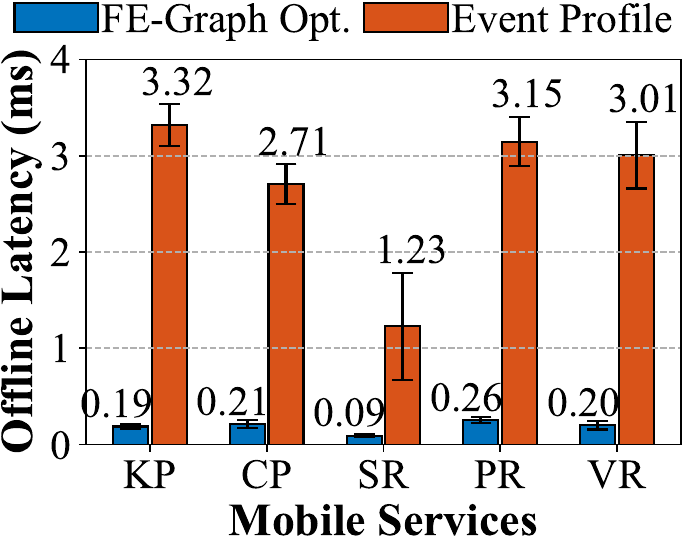}
            \label{fig: offline overhead}
        }
        \subfigure[Online: memory footprint.]{
            \includegraphics[width=\linewidth]{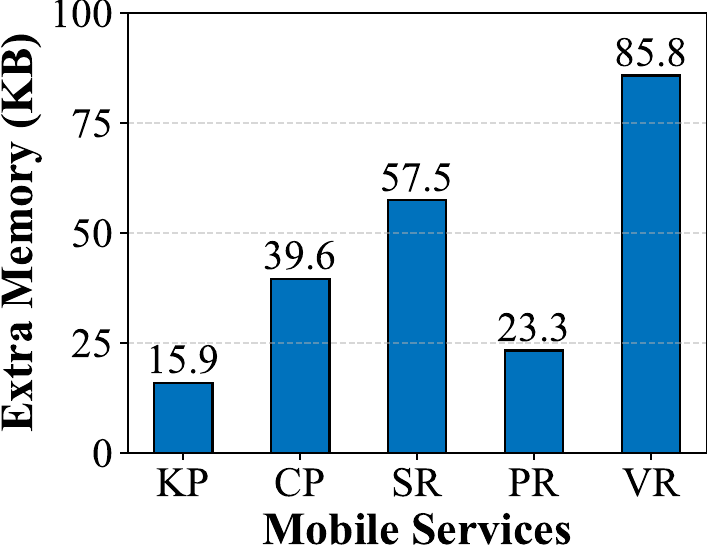}
            \label{fig: online overhead}
        }
        \vspace{-0.3cm}
        \caption{System overheads introduced by AutoFeature.}
        \vspace{-0.3cm}
    \end{minipage}
\end{figure*}

\subsection{Overall Performance}
\label{sec: overall performance}
We begin by evaluating the overall performance of AutoFeature across diverse mobile services.
Figure \ref{fig: overall performance} plots the model execution latencies of different on-device ML models in varied time periods, which are averaged on testing users. 

\textbf{AutoFeature significantly accelerates end-to-end on-device model execution.}
For mobile services in various domains, AutoFeature consistently reduces the on-device model execution latency to below 20ms, which exceeds the human perception range of 30-60 FPS and successfully becomes imperceptible to users.
Compared to the original industry-standard feature extraction process (\textit{w/o AutoFeature}), AutoFeature achieves inference speedups ranging from 1.72-3.44$\times$ for CP service, 1.33-1.44$\times$ for KP service, 1.41-4.53$\times$ for SR service, 1.82-2.18$\times$ for PR service and 3.93-4.43$\times$ for VR service. 
Additionally, we observe that AutoFeature's performance is closely related to the original execution latency (i.e., \textit{w/o AutoFeature}): the largest speedups are observed in CP, SR and VR models with average latency exceeding 40 ms, while the lowest speedup is achieved by KP model with around 20 ms latency.
The reason is that AutoFeature reduces the practical feature extraction latency to relatively low and leaves model inference time as the dominant factor that limits the end-to-end model execution acceleration.

\textbf{The performance of AutoFeature continually improves with time.}
Interestingly, from Figure \ref{fig: overall performance}, we notice that AutoFeature delivers higher speedups of model execution at night compared to daytime periods, which holds across nearly all testing mobile services. 
For example, for CP service, AutoFeature achieves speedups of 1.72$\times$ and 2.79$\times$ during noon and evening, but reaches as high as $3.44\times$ at night.
Upon deeper analysis, we attribute this performance variation to common patterns of user behaviors throughout a day. At night, users tend to engage more actively with mobile apps over an extended and uninterrupted period. This results in a higher volume of newly logged behavior events, increasing the opportunity for AutoFeature to optimize feature extraction by eliminating redundant computations. In contrast, during midday and evening breaks, user interactions with mobile apps are typically shorter and less frequent, due to normal work schedules. 

\textbf{The contributions of redundancy elimination across features and inferences vary across mobile services.}
Figure \ref{fig: overall performance} also reveals that both the graph optimizer (\textit{w/ Fusion}) and cache policy (\textit{w/ Cache}) could accelerate on-device model execution for all mobile services, except for the first model execution during each time period as app exit frees up memory. 
However, their effectiveness varies significantly across different services. While the \textit{w/ Fusion} plays a dominant role in accelerating CP, KP, and PR model executions, it is less effective for SR and VR models. This discrepancy primarily arises from the differences in their overlapping degrees of behavior events across models. 
Specifically, as shown in Figure \ref{fig: feature distribution}, 80.2\% of features in CP, 85\% in KP, and 80.6\% in PR share identical \textit{event\_name} conditions, i.e., the same relevant behavior types, which enables AutoFeature to eliminate a substantial portion of redundant computations across features. In contrast, the SR and VR models exhibit relatively lower overlap (59\% and 71\%, respectively), reducing the optimization potential of inter-feature fusion.

\textbf{AutoFeature introduces marginal extra system costs for both offline and online phases.}
During the offline optimization phase, the primary system overhead comes from 
(i) constructing and optimizing the FE-graph for each model, and 
(ii) profiling the operation cost and result size for each interaction event type.
As shown in Figure \ref{fig: offline overhead}, the overall time cost of offline optimization is dominated by the profiling process, which ranges from 1.23 ms to 3.32 ms across different on-device ML models. Note that the memory consumption during this phase is negligible, as AutoFeature requires only a few kilobytes space to analyze the conditions across features.
During the online execution phase, the extra system overhead stems from caching intermediate results to reduce redundant operations across consecutive inferences. 
As depicted in Figure \ref{fig: online overhead}, the average extra memory footprint required to cache all intermediate results remains consistently below 100 KB. Such low memory cost is generally acceptable for both high-end and low-end mobile devices, attributed to (i) the compact feature sets of lightweight on-device ML models, (ii) our efficient event-level data caching for all features. 
As a result, our proposed greedy cache policy is mainly utilized when the memory budget of each ML model is strictly limited by operating system, analyzed in \S\ref{sec: component-wise analysis}.

\textbf{Comparison with Cloud-based Methods}
    We further benchmark AutoFeature against two cloud-side feature extraction systems. They trade real-time computation for increased storage by offloading expensive operations to an offline logging process and maintaining an additional database to store pre-computed outputs. Since $Decode$ and $Retrieve$ are the top-2 computationally expensive operations, we implemented two baselines based on which operation to offload: Decoded Log and Feature Store. Table \ref{tab: introduction to cloud-side baseline} elaborates their storage structures and offloaded operations.
\begin{table*}[]
    \centering
    \vspace{-0.2cm}
    \caption{
        Detailed introduction to cloud-side feature extraction baselines.}
    \Description{Detailed introduction to cloud-side feature extraction baselines.}
    \label{tab: introduction to cloud-side baseline}
    \vspace{-0.3cm}
     \begin{small}
    \begin{tabular}{|c|c|c|c|}
        \hline
        \textbf{System} & \makecell{\textbf{Offloaded} \textbf{Operations}} & \textbf{Storage Structure} & \textbf{Introduced Storage}\\
        \hline
        \multirow{2}{*}{AutoFeature} & \multirow{2}{*}{\ding{56}} & Each Row: One Behavior Event & \multirow{2}{*}{\ding{56}}\\
        \cline{3-3}
        & & Each Column: Compressed Attributes & \\
        \hline
        \multirow{2}{*}{Decoded Log} & \multirow{2}{*}{Decode}& Each Row: One Behavior Event & \multirow{2}{*}{Massive Columns}\\ 
        \cline{3-3}
         & & Each Column: One Unique Attribute & \\
        \hline
        \multirow{2}{*}{Feature Store} & Decode & Each Row: One Behavior Event required by One Feature & Redundant Rows\\
        \cline{3-3}
         & Retrieve & Each Column: One Unique Attribute & Massive Columns\\
        \hline
    \end{tabular}
\end{small}
    \vspace{-0.3cm}
\end{table*}

\begin{figure*}
    \centering
    \subfigure[Inference Latency.]{
        \includegraphics[height=2.6cm]{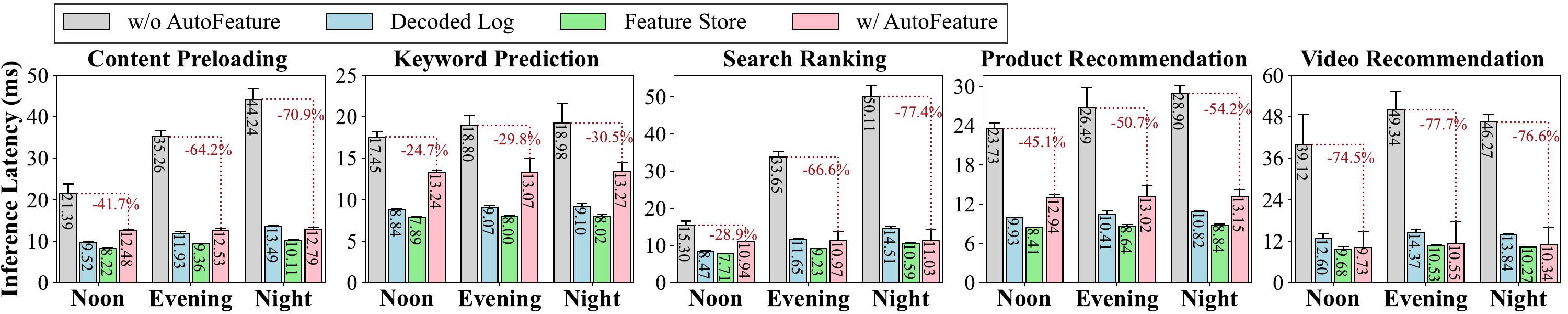}
        \label{fig: compare with cloud performance}
    }
    \ \ 
    \subfigure[Storage Cost.]{
        \includegraphics[height=2.4cm]{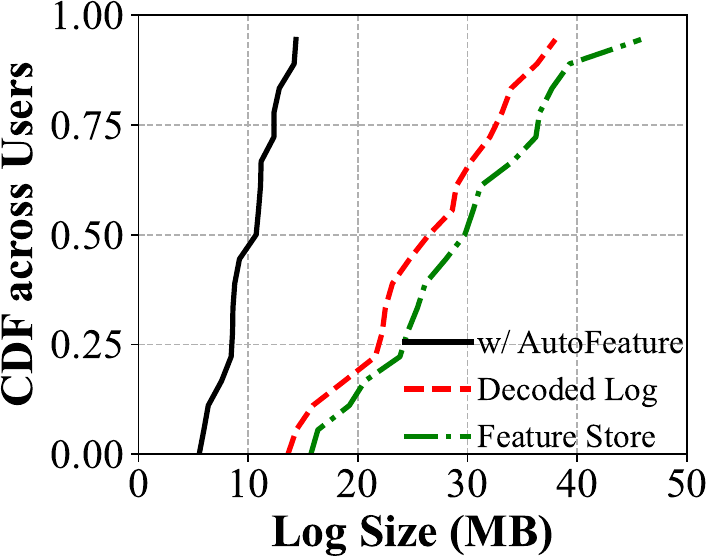}
        \label{fig: compare with cloud cost}
    }
    \vspace{-0.5cm}
    \caption{
        Performance comparison between AutoFeature and cloud-side feature extraction baselines.}
\end{figure*}

    Figure \ref{fig: compare with cloud performance} plots the average inference latency across various mobile services. AutoFeature significantly reduces inference latency across the five services by up to 70.9\%, 30.5\%, 77.4\%, 54.2\% and 77.7\%, and reduces latency by up to 31.35 ms, 5.85 ms, 38.76 ms, 15.68 ms and 38.94 ms. Offloading only $Decode$ (Decoded Log) yields an additional gain of at most $4.38$ ms, and offloading both $Decode$ and $Retrieve$ (Feature Store) further reduces latency by at most $3.91$ ms. However, the cloud baselines accelerate feature extraction with a catastrophic increase in storage requirements, making them impractical for mobile deployment. Figure \ref{fig: compare with cloud cost} illustrates the app log size distributions across testing users for each system. For an average user, Decoded Log increases the app log size by $2.61\times$, and Feature Store increases it by a staggering $2.80\times$. This massive storage inflation is unacceptable for production mobile apps due to its direct link to user churn and financial loss: (i) Public statistics confirm that excessive app size is a primary driver of app uninstallation~\cite{app_uninstall}; (ii) Our internal industrial data reveals that every additional 10 MB in app size leads to a decrease of around 30,000 to 61,000 daily active users, resulting in a daily financial loss of over \$7,000. The storage efficiency provided by AutoFeature is thus a prerequisite for real-world deployment.

\subsection{Component-Wise Analysis}
\label{sec: component-wise analysis}
Next, we evaluate the effectiveness of each key design in AutoFeature. The experiments are mainly conducted on the video recommendation service with the most complex feature dependencies.

\begin{figure*}
    \vspace{-0.2cm}
    \begin{minipage}[b]{0.405\linewidth}
        \centering
        \subfigure[Effect of Inter-Feature Fusion]{
            \includegraphics[height=2.5cm]{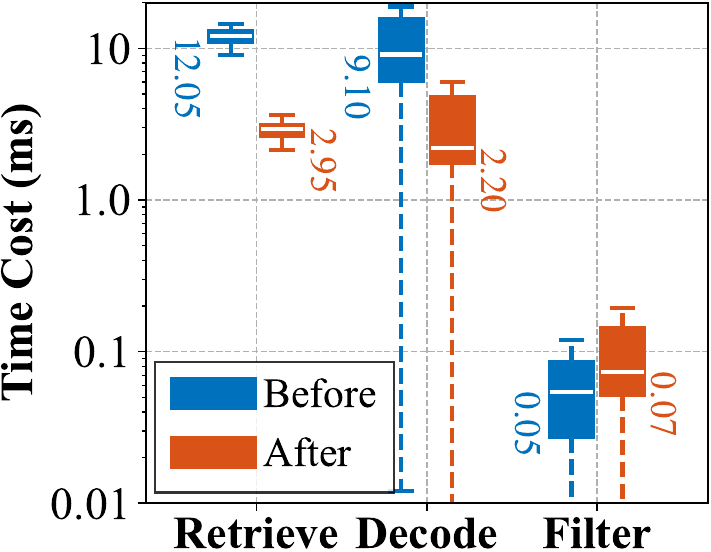}
            \label{fig: component fusion}
        }
        \subfigure[Effect of Cache Policy.]{
            \includegraphics[height=2.5cm]{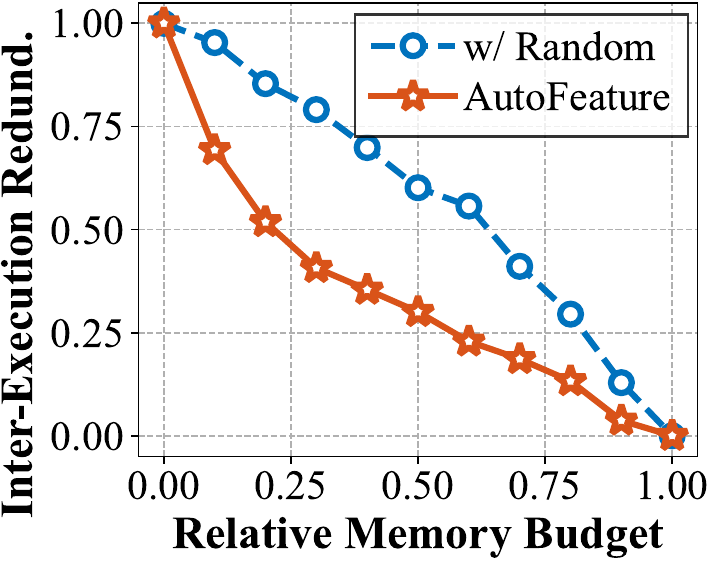}
            \label{fig: component cache}
        }
        \vspace{-0.5cm}
        \caption{Component-wise analysis.}
        \vspace{-0.3cm}
    \end{minipage}
    \ 
    \begin{minipage}[b]{0.245\linewidth}
        \centering
        \includegraphics[height=2.5cm]{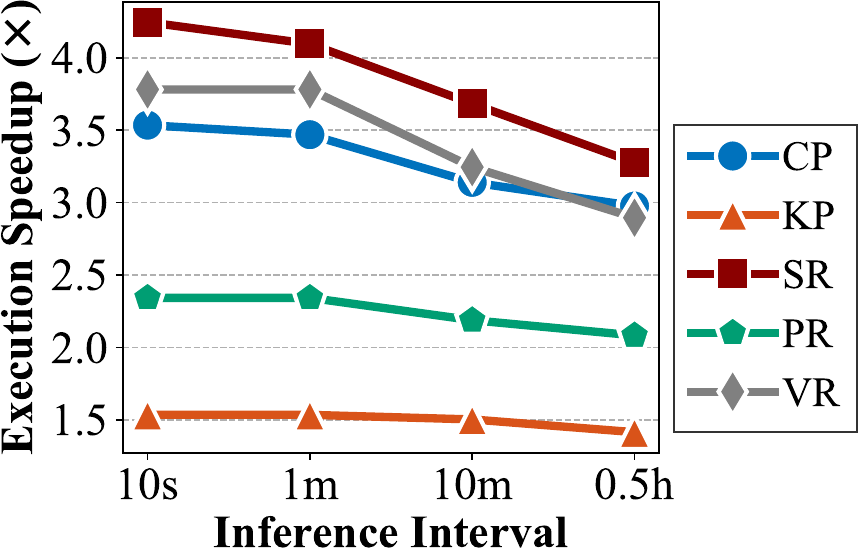}
        \vspace{-0.3cm}
        \caption{Impact of inference interval across services.}
        \vspace{-0.3cm}
        \label{fig: sensitivity to inference interval}
    \end{minipage}
    \ 
    \begin{minipage}[b]{0.33\linewidth}
        \centering
        \includegraphics[height=2.6cm]{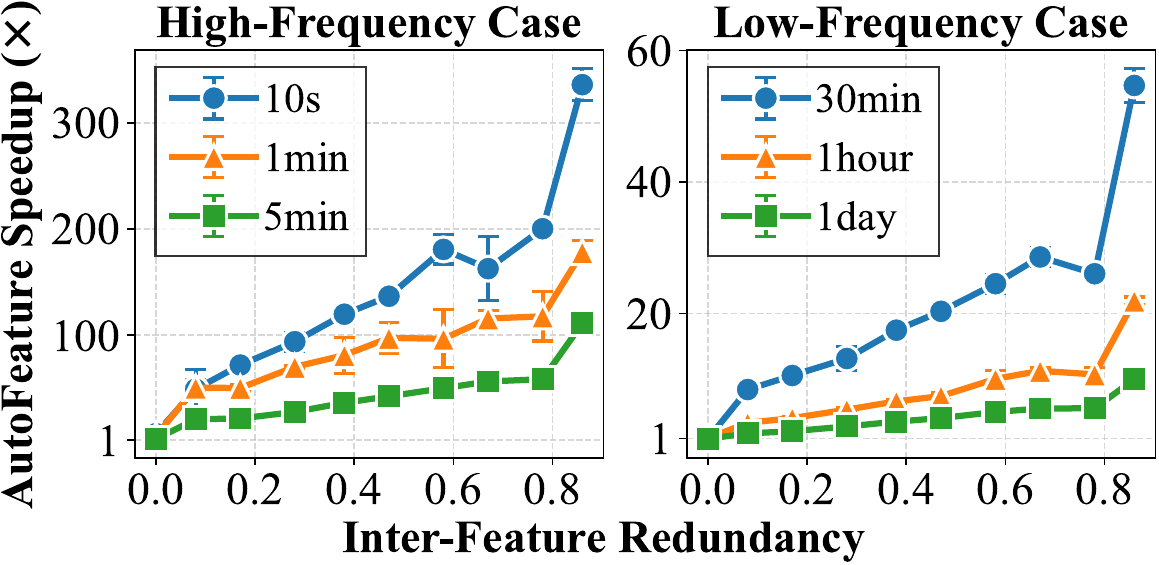}
        \vspace{-0.3cm}
        \caption{Impact of inter-feature redundancy through offline evaluation.}
        \vspace{-0.3cm}
        \label{fig: sensitivity to inter-feature redundancy}
    \end{minipage}
\end{figure*}
\textbf{Inter-Feature Fusion.}
To assess the impact of inter-feature operation fusion, we conduct a breakdown analysis of the feature extraction latency, isolating the time cost of each operation. 
Figure \ref{fig: component fusion} compares the latency distributions across online executions \textit{before} and \textit{after} fusing redundant operations.
Our key finding is that the primary computational savings of inter-feature fusion stem from two bottleneck operations: \textit{Decode} and \textit{Retrieve}. 
Specifically, the average latency of \textit{Decode} decreases from 12.01 ms to 2.95 ms and \textit{Retrieve} decreases from 9.12 ms to 2.23 ms, yielding over 4$\times$ speedups for both of them.
However, the fusion process slightly increases the latency of \textit{Filter} operation. This is in expectation because the fused \textit{Filter} operation has to process and differentiate outputs for each fused feature, introducing extra branching cost. Benefited from our proposed hierarchical filtering algorithm, the extra latency is reduced to only 0.02 ms on average.

\textbf{Inter-Execution Optimization.}
Next, we evaluate the performance of greedy cache policy in reducing redundancy across model executions with various memory constraints.
We compare AutoFeature with a modified version (\textit{w/ Random}), which caches intermediate results of different behavior types randomly rather than utility-to-cost ratio.
Because (i) the redundancy between consecutive inferences varies during online evaluation and (ii) the number of intermediate results differs across mobile users, we analyze the relative changes in redundancy as a function of the proportion of intermediate results cached by the device.
As shown in Figure \ref{fig: component cache}, greedy cache policy consistently outperforms the \textit{w/ Random}, demonstrating its robust performance in handling various memory budgets. 
We also notice that the greedy policy is more effective when device memory is limited, such as reducing 50\% redundant feature extraction operations by caching only 23\% intermediate results. 
This is because the behavior types prioritized by AutoFeature exhibit higher utility-to-cost ratio, implying more redundancy reduction per memory cost unit, highlighting AutoFeature' superior performance when handling strictly limited device memory.

\subsection{Sensitivity Analysis}
\label{sec: micro-benchmarks}
Since AutoFeature operates without hyper-parameters, eliminating the need for trial-and-error tuning, we focus our sensitivity analysis on two environmental factors.

\textbf{Impact of Model Execution Frequency.}
To evaluate how execution frequency affects AutoFeature’s performance, we systematically vary the intervals between consecutive model executions. Specifically, we force on-device ML models of different mobile services to be invoked at fixed intervals during a user's app usage at night.
As shown in Figure \ref{fig: sensitivity to inference interval}, the speedups achieved by AutoFeature decrease as the interval between model executions increases. This trend is expected, as longer intervals reduce the likelihood of overlapping behavior events processed by consecutive feature extractions, thereby limiting the potential for redundancy elimination. 
However, even under infrequent scenarios (e.g., 1 model execution per 30 minutes), AutoFeature still achieves significant speedups ranging from 1.40$\times$ to 2.8$\times$ across real-world mobile services. This result demonstrates AutoFeature’s robustness in handling diverse inference frequencies.

\textbf{Impact of Feature Redundancy.}
To further assess AutoFeature’s performance across potential ML models with varying levels of inter-feature redundancy, we construct numerous synthetic feature sets with controlled redundancy levels. 
Specifically, we define feature redundancy as the proportion of overlapping time ranges among features that rely on the same user behavior types. For each redundancy level, we measure AutoFeature’s speedups on model execution under two scenarios: (i) high-frequency inferences with intervals ranging from 10 seconds to 5 minutes, and (ii) low-frequency inferences with intervals ranging from 30 minutes to 1 day.
As shown in Figure \ref{fig: sensitivity to inter-feature redundancy}, AutoFeature’s performance consistently improves with increasing feature redundancy across various inference frequencies. 
When inference requests occur every 10s and 1h, the feature extraction speedups grow from 7.3$\times$ and $1.0\times$ at 0\% redundancy to 336$\times$ and 21.9$\times$ at nearly 90\% redundancy. Even for the longest inference interval of 1 day, AutoFeature still reduces feature extraction latency by 2.1$\times$, 4.1$\times$ and 5.6$\times$ for redundancy levels of 20\%, 50\% and 80\%.
Te observed speedups do not increase linearly with redundancy levels, as the plotted feature redundancy is estimated based on overlapping time ranges rather than specific behavior events, which are dynamic with user behaviors. 
It is important to note that the speedups reported in Figure \ref{fig: sensitivity to inter-feature redundancy} exceed those observed in online evaluations, as the synthetic experiments isolate and measure only the feature extraction process, whereas online evaluations include the model inference latencies.

%% file: Contents/5-RelatedWork.tex
\section{Discussion on Limitations}
    \textbf{Model-Engine Co-Design.}
    In practice, the ML development pipeline is often divided between two distinct teams: the Model Team, which focuses on maximizing model accuracy and company profit, and the Infrastructure Team, which focuses on optimizing model execution speed. While this decoupling makes our team's feature extraction engine model-agnostic and highly generalizable, it inherently limits the opportunities for model-engine co-design. A co-designed approach could unlock greater efficiency by trading a small, acceptable amount of accuracy for significant latency gains, such as reusing stale feature values rather than recomputing the fresh ones. However, these optimizations are currently hindered by the prevailing organizational silos and technical separation typical of mobile application companies.

\textbf{Dependency on Active Users and App Services.}
    The applicability of AutoFeature is naturally limited to user scenarios and application types where feature extraction constitutes a measurable performance bottleneck. For inactive users with minimal app logs, the feature extraction overhead might be minimal and lower than model inference time, as only a few behaviors are processed. However, this limitation applies only to an edge case, as nearly all major mobile services prioritize optimizing the experience of active users, who are the primary drivers of company revenue and inference requests. 
    Also, for future intensive services where models become larger, the model inference time can dominate the execution pipeline. While this will diminish the proportional impact of our optimization on the total latency, we believe that the need for real-time responsiveness will remain paramount for the main mobile services and make AutoFeature still applicable. 

\section{Related Work}
\label{sec: related work}
\textbf{Device-Side Inference Acceleration.}
Extensive research has been conducted to optimize on-device model inference, which can be classified into four categories: 
\textit{operator optimization}~\cite{kong2023convrelu++, jiang2020mnn, niu2021dnnfusion} to enhance backend inference engines at the operator level, 
\textit{model architecture optimization} through quantization~\cite{liu2018demand, kim2019mulayer, lin2024awq}, pruning~\cite{DBLP:conf/mobicom/WenLZJYOZL23, shen2024fedconv, ma2020pconv, niu2020patdnn}, sparsification~\cite{lym2019prunetrain, bhattacharya2016sparsification} and architecture re-design~\cite{tang2023lut, guo2021mistify, chen2020deep} to reduce computational complexity, 
\textit{hardware resource exploitation}~\cite{khani2023recl, jeong2022band, wang2021asymo, DBLP:conf/mobisys/CaoBB17, DBLP:conf/mobicom/XuXWW0H0JL22, jia2022codl} to utilize multiple types of on-device computational hardware to accelerate computation, and
\textit{inference frequency reduction}~\cite{yuan2022infi, li2020reducto, chen2015glimpse, guo2018foggycache} to avoid unnecessary inference requests.
However, these work primarily focused on optimizing the model execution stage due to exclusively considering traditional vision or language models that use static input features.
As a result, AutoFeature is complementary to them by optimizing the previous feature extraction stage for practical on-device ML adoption in mobile apps.

\textbf{Cloud-Side Feature Computation.}
Another related research topic is feature computation at cloud servers. 
To provide comprehensive features for real-time model inference processes requested by different online services, many service providers maintain an up-to-date feature store at cloud servers like 
Amazon SageMaker~\cite{amazon_feature_store} and Databricks Feature Store~\cite{databricks_feature_store}
, which collects all user data, pre-computes and stores necessary user features. This setups allows multiple application services to concurrently query user features~\cite{wooders2023ralf, kakantousis2019horizontally}, but relies on huge storage and computational capabilities of cloud servers and introduces significant privacy concerns~\cite{voigt2017eu, mazeh2020personal}.
Consequently, there is a growing trend towards offloading both feature extraction and model execution processes to device for more real-time, secure and personalized serves~\cite{gong2025optimizing, DBLP:conf/www/XuLLLLL19}. Our work analyzes the practical issues in on-device feature extraction process and further explores potential optimization directions. 

\textbf{Feature Selection} is a mature field focused on identifying the optimal subset of features to maximize accuracy and minimize dimensionality~\cite{zha2025data, li2017feature}. To achieve this, filter methods valuate features using a scoring function based on statistical properties like mutual information gain~\cite{gao2016variational, shishkin2016efficient, thaseen2017intrusion, du2013local} and representativeness to data distribution~\cite{farahat2011efficient, gu2012generalized, he2005laplacian} . 
    Wrapper methods leverage the model performance to assess the quality of selected features and refining the selection iteratively~\cite{zhang2019active, schnapp2021active, arai2016unsupervised}. However, feature selection operates before deployment to determine what inputs the model should use, and our work operates in the post-deployment stage to optimize how those essential features are extracted for real-time on-device model inferences

%% file: Contents/6-Conclusion.tex
\section{Conclusion}
In this work, we identify an overlooked bottleneck of feature extraction in on-device model inference with user behavior sequences, which are prevalent in real-world industrial mobile apps. 
To address this bottleneck, we propose AutoFeature, the first feature extraction engine designed to accelerate on-device execution by eliminating redundant operations across both input features and consecutive model executions without compromising model accuracy.
We implement a system prototype and integrate it into five real-world mobile services for evaluation, where AutoFeature achieves 1.33$\times$-4.53$\times$ speedup in end-to-end model execution latency.

\begin{acks}
This work was supported in part by National Key R\&D Program of China (No. 2023YFB4502400), in part by China NSF grant No. 62322206, 62432007, 62441236, 62132018, 62025204, U2268204, 62272307, 62372296, U25A6024, in part by Fundamental and Interdisciplinary Disciplines Breakthrough Plan of the Ministry of Education of China (No. JYB2025XDXM103). The opinions, findings, conclusions, and recommendations expressed in this paper are those of the authors and do not necessarily reflect the views of the funding agencies or the government.
Zhenzhe Zheng is the corresponding author.
\end{acks}

%% file: Contents/7-Appendix.tex
\clearpage
\appendix
\section{Supplementary Dataset Analysis}
\label{sec: appendix}
\setcounter{table}{0}
\setcounter{figure}{0}

    In this section, we include supplementary material to provide comprehensive context for our industrial dataset, detailing the specific behavior traces of different video types across testing users. 
    
    \textbf{Diverse Video Behavior Traces.}
    In the Evaluation Setup section, we detailed the overall behavior frequencies per 10-minute segment. Here, we break down those numbers by specific video type. In modern multimedia platforms like TikTok, behaviors such as watching normal short-form videos, live streams, shows, and creator homepage videos are all recorded as distinct behavior types. This distinction is necessary because these types involve heterogeneous attribute set for description and target different audience. 

\begin{figure}[H]
    \centering
    \subfigure[Short-Form Video]{
		\includegraphics[width=0.47\linewidth]{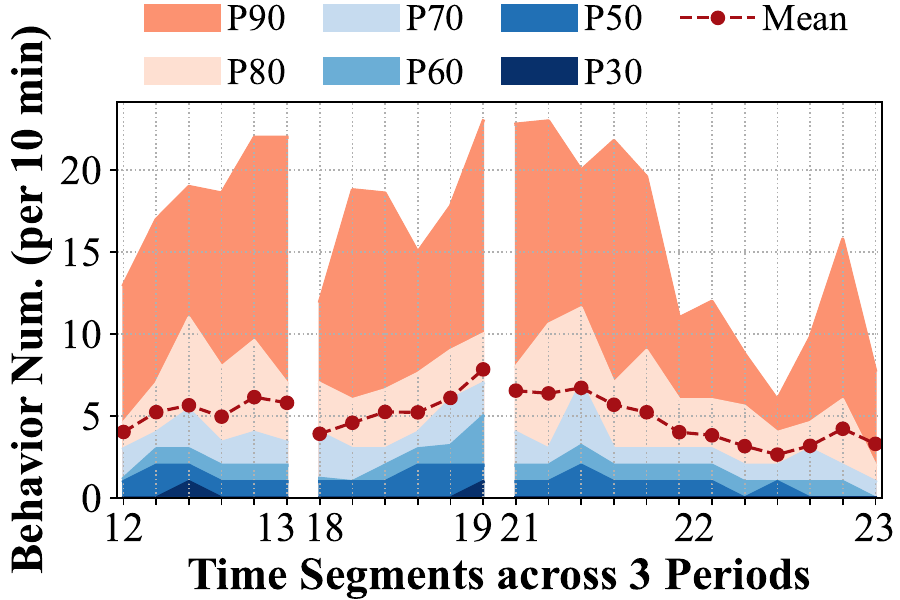}
		\label{fig: traces of video}
	}
	\subfigure[Live Stream]{
		\includegraphics[width=0.47\linewidth]{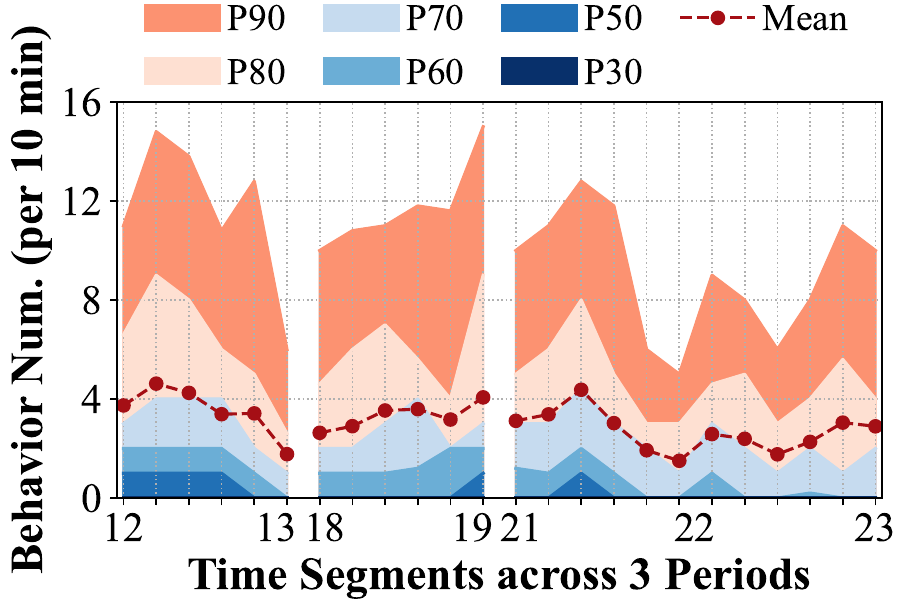}
		\label{fig: traces of live stream}
	}
	\subfigure[Show]{
		\includegraphics[width=0.47\linewidth]{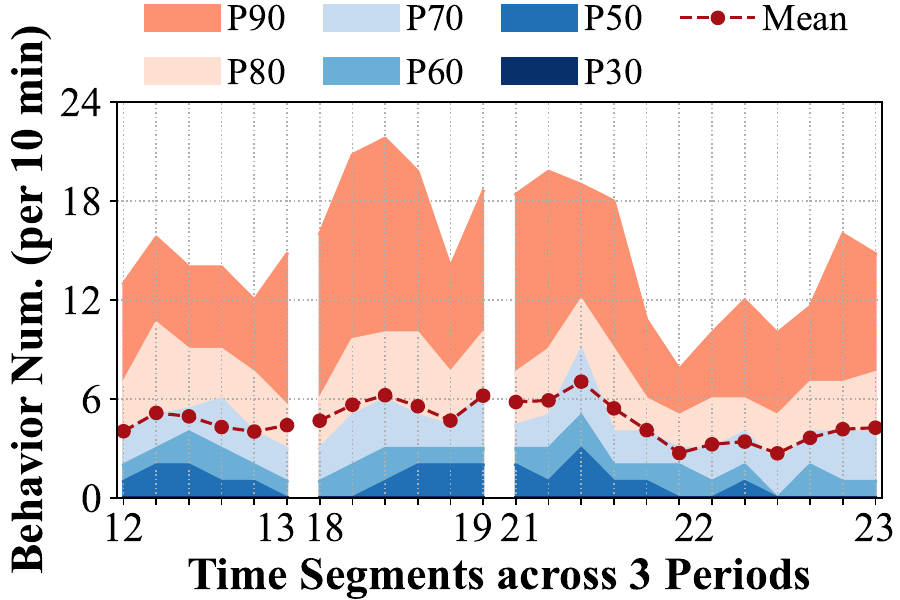}
		\label{fig: traces of show}
	}
	\subfigure[Homepage]{
		\includegraphics[width=0.47\linewidth]{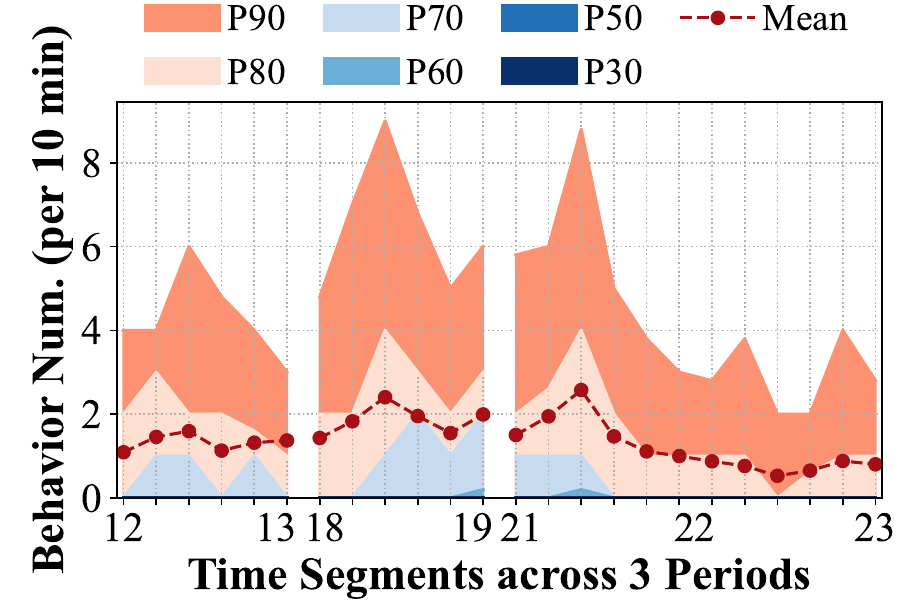}
		\label{fig: traces of homepage}
	}
    \vspace{-0.3cm}
	\caption{Behavior traces of testing users across three time periods (noon, evening and night). Each subfigure focuses on one video-related behavior and we plot the frequencies within each 10-minute segment of users with different activity levels (P90, P80, P70, P60, P50, and P30 traces).}
    \vspace{-0.3cm}
	\label{fig: behavior traces}
\end{figure}

    Figure \ref{fig: behavior traces} plot the frequencies of these distinct behaviors per 10-minute segment across our testing users. 
    For the most popular short-form videos as shown in Figure \ref{fig: traces of video}, testing users consume an average of 4.02–6.15 short-form videos per 10 minutes at noon, 3.91–7.84 at evening, and 2.64–6.71 at night. Notably, the consumption rate is slightly lower at night, which may be explained by users dedicating more short, intense bursts of leisure time (like noon breaks) to this format. 
    Shows and live streams are the next two most popular types and as shown in Figures \ref{fig: traces of live stream} and \ref{fig: traces of show}, average user consumptions per 10 minutes are 1.77-4.62 live stream and 4.04-5.15 shows at noon, 2.63-4.06 and 4.68-6.22 for evening as well as 1.50-4.37 and 2.72-7.05 for night. 
    The lowest frequency behavior is viewing a creator's homepage (Figure \ref{fig: traces of homepage}), averaging only 0.52–2.40 views per 10 minutes across all three time periods.
    It is important to note that videos of different genres (e.g., comedy, drama) are typically recorded under the same behavior type. This is because they share the same underlying set of descriptive attributes, differing only in the values of those attributes.
